# scientific reports

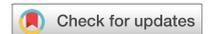

# OPEN   An arithmetic method algorithm optimizing k-nearest neighbors compared to regression algorithms and evaluated on real world data sources

Theodoros Anagnostopoulos[1✉], Evanthia Zervoudi[1,4], Christos Anagnostopoulos[2,4], Apostolos Christopoulos[1,4] & Bogdan Wierzbinski[3,4]

Linear regression analysis focuses on predicting a numeric regressand value based on certain regressor values. In this context, k-Nearest Neighbors (k-NN) is a common non-parametric regression algorithm, which achieves efficient performance when compared with other algorithms in literature. In this research effort an optimization of the k-NN algorithm is proposed by exploiting the potentiality of an introduced arithmetic method, which can provide solutions for linear equations involving an arbitrary number of real variables. Specifically, an Arithmetic Method Algorithm (AMA) is adopted to assess the efficiency of the introduced arithmetic method, while an Arithmetic Method Regression (AMR) algorithm is proposed as an optimization of k-NN adopting the potentiality of AMA. Such algorithm is compared with other regression algorithms, according to an introduced optimal inference decision rule, and evaluated on certain real world data sources, which are publicly available. Results are promising since the proposed AMR algorithm has comparable performance with the other algorithms, while in most cases it achieves better performance than the k-NN. The output results indicate that introduced AMR is an optimization of k-NN.

## Problem definition

Regression analysis focuses on predicting a regressand numeric value based on a certain vector of regressor values. Specifically, when there is considered a linear relation between the regressors' and the regressand values focus is given on linear regression. Concretely, there are intelligent algorithms used to solve linear regression problems, while optimization of the existing regression algorithms is an area of continuous research direction. Although k-Nearest Neighbors (k-NN) regression is a fundamental non-parametric algorithm used both on regression and classification problems there are cases where its efficiency could be further optimized. In this research effort, an arithmetic method is introduced to solve linear equations with an arbitrary number of real variables, which is also supported by the proposed Arithmetic Method Algorithm (AMA). Intuitively, such variables are considered as possible regressors feeding a linear regression problem thus optimizing existing structure of k-NN regression algorithm. The outcome of this combination is an optimized k-NN algorithm introduced as Arithmetic Method Regression (AMR) algorithm. Consequently, AMR is evaluated on real-world data sources based on certain methods and metrics, such as leave-one-out cross-validation, mean absolute error (MAE), mean squared error (MSE), and R-squared ($R^2$) as the coefficient of determination. Additionally, time complexity is also evaluated by measuring execution time (ET), since it is also a performance metric defining the efficiency of an intelligent algorithm in certain areas of interest. Subsequently, proposed algorithm is compared with other regression algorithms in literature to assess its result performance. Intuitively, statistical significance of the predicted regression result values is evaluated with a two-tailed permutation test (i.e., randomization test) for each of the adopted machine learning algorithms. Such intelligent algorithms are the k-NN, Linear Regression (LR), Decision Tree (DT), Support Vector Regression (SVR) Random Forest (RF), eXtreme Gradient Boosting (XGBoost), and Convolutional Neural Network (CNN) regression algorithms, while all algorithms are evaluated on the same real world data sources.

[1]Department of Business Administration, University of the Aegean, Chios Island, North Aegean, Michalon 8, Chios 821 00, Greece. [2]School of Computing Science, University of Glasgow, Glasgow G12 8QQ, Sir Alwyn Building, Office S114, UK. [3]Faculty of Economics and Finance, University of Rzeszow, Office D1 3, ul. Cwiklinska 2, Rzeszow 35–611, Poland. [4]These authors contributed equally to this work: Evanthia Zervoudi, Christos Anagnostopoulos, Apostolos Christopoulos and Bogdan Wierzbinski. ✉email: Theodoros.Anagnostopoulos@aegean.gr

## Related work

The problem of regression analysis is treated widely in research literature by various regression algorithms. However, in the context of the current research effort focus is given on comparing several regression algorithms with the k-NN algorithm or other optimized variations of the k-NN algorithm thus providing a basis for comparison between the evaluated algorithms. Concretely, this is a conditional assumption towards performing comparative assessment between the adopted regression algorithms and the proposed AMR algorithm.

Specifically, regression analysis focuses on forensic and anthropological data sources in the research area of dental age estimation, where the adopted regression experiments investigate the prediction performance of random forest (RF), support vector machine (SVM), k-NN, gradient boosting method, LR, and segmented normal Bayesian calibration algorithms[1]. Distance function of the k-NN algorithm is the Euclidean distance. Comparison performed with regards to the MAE, RMSE, Inter-Quartile Range of the error distribution and the slope of the estimated age error, where most of the adopted algorithms demonstrate low levels of error but also exhibited significant bias. Overall, segmented normal Bayesian calibration algorithm showed no significant bias thus it achieved optimal performance than the other algorithms. Concretely, regression analysis focuses on biological age and health check-up data sources in the research area of increased and healthy life expectancy, where the adopted regression experiments investigate the prediction performance of LR, elastic net, least absolute shrinkage and selection operator (LASSO) regression, ridge regression, RF, SVM, gradient boost, and k-NN algorithms[2]. Distance function of the k-NN algorithm is the Euclidean distance. Comparison performed with regards to the MSE and the $R^2$, where SVM achieved optimal performance with lower value of error and higher value of $R^2$. Overall, optimal results were observed by the SVM algorithm. Intuitively, regression analysis focuses on glucose measurement of hospitalized patients' data sources in the research area of continuous glucose monitoring and outpatient blood glucose control, where the adopted regression experiments investigate the prediction performance of time series analysis, cubist-rule based regression, LR, RF, partial least squares, and k-NN algorithms[3]. Distance function of the k-NN algorithm is the Euclidean distance. Comparison performed with regards to the RMSE, $R^2$ and Clarke error grid, where k-NN resulted to the worst performance than the other algorithms. Overall, optimal performance is achieved with cubist-rule based regression. Subsequently, regression analysis focuses on cuckoo search-based metaheuristic approach data sources in the research area of patients' depression, where the adopted regression experiments investigate the prediction performance of convolutional neural network-enhanced with particle swarm-cuckoo search, k-NN, SVR, DT, residual neural network, visual geometry group, neural networks, logistic regression, and ridge regression algorithms[4]. Distance function of the k-NN algorithm is the Euclidean distance. Comparison performed with regards to the MSE, and $R^2$ indicates that the convolutional neural network-enhanced particle swarm-cuckoo search outperformed other algorithms, thus k-NN, DT, logistic regression and ridge regression achieved lower performance. Overall, optimal performance is achieved with the convolutional neural network-enhanced with particle swarm-cuckoo search.

Continuously, regression analysis focuses on climate and air pollution data sources in the research area of asthma hospitalizations, where the adopted regression experiments investigate the prediction performance of RF, eXtreme Gradient Boosting (XGBoost), multiple linear regression, SVM, and k-NN algorithms[5]. Distance function of the k-NN algorithm is the Euclidean distance. Comparison performed with regards to the MSE, RMSE and $R^2$, resulting in that the RF has better performance than the k-NN and the other algorithms. Overall, optimal performance is observed with the RF algorithm. Concretely, regression analysis focuses on heart rates accelerometer data sources in the research area of patients' heart diseases, where adopted regression experiments investigate the prediction performance of autoregressive integrated moving average (ARIMA)-enhanced with LR, SVR, k-NN, DT, RF, and long short-term memory (LSTM) neural network algorithms[6]. Distance function of the k-NN algorithm is the Euclidean distance. Comparison performed with regards to the MSE, RMSE, and scatter index (SI), where ARIMA-enhanced with LR results to efficient results when validated with walk-forward accelerometer data than other more complex algorithms such as the LSTM neural network. Overall, optimal performance is achieved with ARIMA-enhanced with LR. Intuitively, regression analysis focuses on air pollution, tobacco use, socioeconomical status, employment status, marital status and environment data sources in the research area of lung cancer incidence rate, where adopted regression experiments investigate the prediction performance of LR, SVR, RF, k-NN, and cubist tree algorithms[7]. Distance function of the k-NN algorithm is the Euclidean distance. Comparison performed with regards to the RMSE, and $R^2$, where RF without feature selection as well as with feature selection could support the interpretation of the most contributing variables that the other regression algorithms. Overall, optimal performance is observed with the RF algorithm. Subsequently, regression analysis focuses on life satisfaction data sources in the research area of older people real-life relationships, where adopted regression experiments investigate the prediction performance of SVR, multiple linear regression, ridge regression, LASSO, k-NN, and DT algorithms[8]. Distance function of the k-NN algorithm is the Euclidean distance. Comparison performed with regards to the MAE, MSE, RMSE and $R^2$, result that although k-NN and DT exhibited the best model fitting ability have instead indicated the poorest algorithm validation and algorithm generalization factors than the SVR. Overall, optimal performance is achieved with the SVR algorithm.

Intuitively, regression analysis focuses on epigenetic age acceleration data sources in the research area of stroke patients' vascular risk factors, where adopted regression experiments investigate the prediction performance of elastic net regression, k-NN, RF, SVR, and multilayer perceptron algorithms[9]. Distance function of the k-NN algorithm is the Euclidean distance. Comparison performed with regards to the MAE, RMSE, and $R^2$, where best results are achieved by elastic net regression and multilayer perceptron than the other algorithms including the k-NN algorithm. Overall, optimal performance is observed by both the elastic net regression as well as the multilayer perceptron algorithms on the provided data sources. Continuously, regression analysis focuses on high blood pressure and health complications data sources in the research area of patients' cuffless blood pressure monitoring, where adopted regression experiments investigate the prediction performance of tree-based pipeline optimization tool (TPOT), RF, and k-NN algorithms[10]. Distance function of the k-NN algorithm is the Euclidean distance. Comparison performed with regards to the MAE, MSE, and $R^2$ where TPOT outperformed the other regression algorithms on the provided data sources. Overall, the proposed TPOT algorithm has achieved the optimal performance. Concretely, regression analysis focuses on the application of intelligent regression algorithms for predicting patients' survival in cases of ovarian cancer, where adopted regression experiments investigate the prediction performance of k-NN, SVM, DT, RF, adaptive boosting (AdaBoost), and XGBoost algorithms[11]. Distance function of the k-NN algorithm is the Euclidean distance. Comparison performed with regards to the RMSE, $R^2$ and Pearson's correlation coefficient indicates that the XGBoost is more efficient than the other regression algorithms including RF for the given data sources. Overall, optimal performance is observed by the XGBoost regression algorithm. Additionally, regression analysis focuses on nuclear magnetic resonance spectroscopy data sources in the research area of biological metabolomic age clocks of participants' health and life span, where adopted regression experiments investigate the prediction performance of ridge regression, LASSO, elastic net, partial least squares regression, multivariate adaptive regression splines, SVR, DT, bagging, RF, XGBoost, Bayesian additive regression trees, cubist-rule based regression, and k-NN algorithms[12]. Distance function of the k-NN algorithm is the Minkowski distance. Comparison performed with regards to the MAE, RMSE, Pearson's correlation coefficient, and $R^2$, indicate that the k-NN algorithm performed overfitting on the provided data sources compared with the other algorithms. Overall, optimal performance is achieved with cubist-rule based regression.

Subsequently, regression analysis focuses on the application of intelligent regression algorithms for predicting the reliability of lithium-ion (Li-ion) rechargeable batteries, which are used in various implantable medical devices, where adopted regression experiments investigate the prediction performance of k-NN, and particle swarm optimization algorithm enhanced with Gaussian (Radial Basis Function) RBF kernel[13]. Distance function of the k-NN algorithm is the Euclidean distance. Comparison performed with regards to the RMSE, and maximum error (ME) where k-NN indicates more efficient results compared with the other algorithms. Overall, optimal performance is observed with k-NN algorithm. Concretely, regression analysis focuses on neurological data sources in the research area of segmental motor outcomes in traumatic spinal cord injury, where adopted regression experiments investigate the prediction performance of LR, ridge regression, logistic regression, and k-NN algorithms[14]. Distance function of the k-NN algorithm is the Hamming distance. Comparison performed with regards to the RMSE, where k-NN has the lower value of error compared with the other algorithms. Overall, optimal performance is achieved with k-NN algorithm. Intuitively, regression analysis focuses on human body ground reaction force data sources in the research area of athletes' movement during jumping and inertial sensor measurements, where adopted regression experiments investigate the prediction performance of singular value decompositions (SVD) embedded regression, LSTM Neural Networks, and k-NN algorithms[15]. Distance function of the k-NN algorithm is the Euclidean distance. Comparison performed with regards to the RMSE, and the relative root mean squared error (RRMSE), where k-NN is proved that it can be similarly accurate or more accurate than LSTM neural networks, requiring fewer computing resources and energy, while allowing much faster training times and hyperparameter optimization. Overall, optimal performance is achieved with k-NN algorithm. Continuously, regression analysis focuses on human health status data sources in the research area of preventive healthcare and immediate patients' treatment, where adopted regression experiments investigate the prediction performance of LR, deep neural network, random forest, XGBoost, and k-NN algorithms[16]. Distance function of the k-NN algorithm is the Mahalanobis distance. Comparison performed with regards to the MAE, RMSE, and $R^2$, where k-NN results are significantly improved concerning regression results with the other algorithms. Overall, Optimal performance is observed with k-NN algorithm.

Intuitively, regression analysis focuses on medical data sources in the research area of predicting values of temperature, blood pressure, and cholesterol level with regards to the patient's age, where adopted regression experiments investigate the prediction performance of k-NN, and an affine invariant k-NN estimate algorithms[17]. Distance function of the k-NN algorithm is a proposed affine invariant distance. Comparison performed with regards to the MAE, where affine invariant k-NN estimate has the lower value of error compared with k-NN algorithm. Overall, optimal performance is achieved with affine invariant k-NN estimate algorithm. Additionally, regression analysis focuses on analyzing prediction performance of medical context provided by a Parkinson's disease data source to estimate the infection severity of certain patients, where adopted regression experiments investigate the prediction performance of LR, M5 Prime (M5P) regression tree, neural network, k-NN, mutual k-NN (MkNN), error-based weighting of instances for k-NN (EWkNN), and error correction-based k-NN (ECkNN) regression algorithms[18]. Distance function of the k-NN algorithm is the Euclidean distance. Comparison performed with regards to the MAE and normalized mean absolute error (NMAE), where adopted algorithms demonstrate low levels of error in cases of noiseless input data. However, evaluating the algorithms with data containing additional Gaussian noise results that ECkNN model outperformed the other algorithms. Overall, optimal performance is observed with ECkNN algorithm. Concretely, regression analysis focuses on human health status data sources in the research area of fluoroscopic imaging, which captures X-ray images at video framerates for providing support to catheter insertions assisting vascular surgeons and interventional radiologists, where adopted regression experiments investigate the prediction performance of maximum likelihood estimation, k-NN, smoothed k-NN and k-NN with interior operation parameters (kNN + IOP) algorithms[19]. Distance function of the k-NN algorithm is the Euclidean distance. Comparison performed with regards to the RMSE results that the smoothed k-NN algorithm outperforms k-NN and kNN + IOP regression algorithms. Overall, optimal performance is observed with smooth k-NN. Specifically, regression analysis focuses on health outcomes observed by studying body mass index and cholesterol levels in relation with patient age to estimate associated risk factors., where adopted regression experiments investigate the prediction performance of B spline, k-NN fused LASSO, and alternating direction method of multipliers (ADMM) algorithms[20]. Distance function of the k-NN fussed LASSO algorithm is the Euclidean distance. Comparison performed with regards to the MSE, mean squared difference (MSD), where k-NN fused LASSO has efficient results compared with the other algorithms. Overall, optimal performance is achieved with k-NN fused LASSO. Subsequently, regression analysis focuses on quantitative structure-activity relationship (QSAR) modeling data sources in the research area of drug discovery in molecular design, where adopted regression experiments investigate the prediction performance of RF, graph convolutional neural network, transformer convolutional neural network, transformer convolutional neural network with augmentation, topological regression, topological regression with disjoint anchor and training set, ensemble topological regression, and k-NN enhanced with metric learning for kernel regression algorithms[21]. Distance function of the k-NN algorithm is the Mahalanobis distance. Comparison performed with regards to the normalized root mean squared error (NRMSE), and Spearman's correlation coefficient, where k-NN enhanced with metric learning for kernel regression achieves the best performance results in comparatively high performance compared with the other algorithms. Overall, optimal performance is achieved with k-NN enhanced with metric learning for kernel regression algorithm.

Research in literature indicates that there are simple algorithms (e.g., such as k-NN), which outperform complex algorithms (e.g., such as LSTM neural network) regarding certain evaluation methods and metrics when applied to provided data sources. Intuitively, specific focus is given on k-NN algorithm exploration and further exploitation. Concretely, it can be observed that certain intervention applied on the k-NN algorithm is possible to provide optimized prediction results. Subsequently, focusing on the k-NN linear regression algorithm adoption of the proposed arithmetic method as an additional computational dimension results to AMR algorithm for regression analysis, which is an optimization of k-NN with evidence provided in the current research effort. The rest of the paper is structured as follows. In section methods, the proposed arithmetic method is presented in the form of a theorem, which is proved. Numerical validation of the method is provided and supported by AMA. Consecutively, the proposed AMR algorithm is presented. In results section, real world data sources used in the current research effort are provided, while evaluation methods and metrics are presented. Additionally, it is performed comparison of AMR with other regression algorithms in the literature. In discussion section, the results of proposed algorithmic behavior are analyzed. Concretely, paper concludes by proposing future work directions, which are provided for further research in the domain.

## Methods
### Mathematical foundations of AMA
In this section, we explain how the Arithmetic Method Algorithm (AMA) can be represented in standard linear-algebra form, derive the main inequalities, and discuss its stability, its relation to the Moore–Penrose pseudo-inverse, and its role in the Arithmetic Method Regression (AMR) ensemble. All proofs are constructive and compatible with the algorithmic steps implemented in this study.

### Linear arithmetic method algorithm (AMA)
Let $A \in R^{m \times n}$, $b \in R^m$. AMA seeks an approximate solution $x_{\{AMA\}} A \in R^n$ resulting from:

$$A x = b$$

We assume AMA is linear in b for fixed A : $\exists$ operator $L_{\{AMA\}} \in R^{n \times m}$ such that:

$$x_{\{AMA\}} = L_{\{AMA\}} b$$

### Theorem 1 (existence, residual bound, and stability)
Let A has a full column rank and $x_{LS} = A^+ b$ be the least-squares (Moore–Penrose) solution, with $A^+$ the Moore–Penrose pseudo-inverse.

### Existence and uniqueness
AMA coincides with the LS solution whenever $L_{\{AMA\}} A = I_n$ and in that case:

$$x_{\{AMA\}} = x_{\{LS\}} = A^+ b$$

**Proof** multiplying both sides of $x_{\{AMA\}} = L_{\{AMA\}} b$ by A $\to$ $A x_{\{AMA\}} = A L_{\{AMA\}} b = b$. If A is full column rank, this equation admits the unique solution $x_{\{LS\}} = A^+ b$. Hence AMA produces the exact solution if its implicit left-inverse $L_{\{AMA\}} = A^+$.

### Residual bound

$$\| A x\{AMA\} - b \|2 \le \| A \|2 \cdot \| x\{AMA\} - x\{LS\} \|2 + \| A x\{LS\} - b \|2$$

**Proof** Subtract the defining equations for $x_{\{AMA\}}$ and $x_{\{LS\}}$:

$$A\left(x_{\{AMA\}} - x_{\{LS\}}\right) = \left(A\,x_{\{AMA\}} - b\right) - \left(A\,x_{\{LS\}} - b\right).$$

Taking Euclidean norms and using the triangle inequality gives

$$\|A\,x_{\{AMA\}} - b\|_2 \leq \|A\|_2 \|x_{\{AMA\}} - x_{\{LS\}}\|_2 + \|A\,x_{\{LS\}} - b\|_2.$$

This shows that the AMA residual is bounded by the deviation of $x_{\{AMA\}}$ from the least-squares solution and the standard residual of the pseudoinverse estimator. The term $\|A\,x_{\{LS\}} - b\|_2$ vanishes when the system is consistent.

### Stability under perturbations

For perturbations $(\Delta A, \Delta b)$ satisfying $\|\Delta A\|_2 \leq \eta_A, \|\Delta b\|_2 \leq \eta_b$, there exists C > 0 (depending on $\|A^+\|_2$ and $\|L_{\{AMA\}}\|_2$) such that:

$$\|x_{\{AMA\}}(A + \Delta A, b + \Delta b) - x_{\{AMA\}}(A, b)\|_2 \leq C(\eta_A + \eta_b)$$

**Proof** If $L_{\{AMA\}}A = I_n$, then $Ax_{\{AMA\}} = AL_{\{AMA\}}b = b$; hence $\boldsymbol{x_{\{AMA\}}}$ **solves the system**.

**Uniqueness** follows from the full-rank assumption. When the matrix $A \in R^{m \times n}$ has full column rank (rank(A) = n), its columns are linearly independent. This implies that the normal matrix $A^TA$ is invertible because $A^TAu = 0 \Rightarrow u^TA^TAu = 0 \Rightarrow u = 0$. Hence, the least-squares solution

$$x_{\{LS\}} = \left(A^T A\right)^{-1} A^T b$$

is **unique**.

If two vectors $x_1$ and $x_2$ satisfy $Ax_1 = Ax_2 = b$ then $A(x_1 - x_2) = 0$ which implies $(x_1 - x_2) \in \ker(A)$.

But since the kernel of A is trivial when A is full column rank (ker(A) = {0}), it follows that $x_1 = x_2$. Therefore, the solution is unique, and AMA can at most approximate one single valid vector $x^*$.

Subtract $Ax_{\{LS\}} = b$ to obtain $A\left(x_{\{AMA\}} - x_{\{LS\}}\right) = \left(A\,x_{\{AMA\}} - b\right) - \left(A\,x_{\{LS\}} - b\right)$

The triangle inequality and the operator norm may lead to the **residual bound**.

$$A\left(x_{\{AMA\}} - x_{\{LS\}}\right) = \left(A\,x_{\{AMA\}} - b\right) - \left(A\,x_{\{LS\}} - b\right).$$

Taking the Euclidean ($\ell_2$) norm on both sides:

$$\|A\left(x_{\{AMA\}} - x_{\{LS\}}\right)\|_2 = \|\left(A\,x_{\{AMA\}} - b\right) - \left(A\,x_{\{LS\}} - b\right)\|_2.$$

By the triangle inequality for vector norms,

$$\|A\left(x_{\{AMA\}} - x_{\{LS\}}\right)\|_2 = \|\left(A\,x_{\{AMA\}} - b\right)\|_2 - \|\left(A\,x_{\{LS\}} - b\right)\|_2.$$

Apply the operator norm inequality (or submultiplicative property)[22,23]:

$$\|A\left(x_{\{AMA\}} - x_{\{LS\}}\right)\|_2 = \|A\|_2 \|\left(x_{\{AMA\}} - x_{\{LS\}}\right)\|_2.$$

Combining both inequalities gives.

This is the **residual bound**, which relates the AMA residual to the least-squares residual and the deviation between the two estimates.

Since $x_{\{AMA\}} = L_{\{AMA\}}b$, the total differential gives $dx_{\{AMA\}} = L_{\{AMA\}}db + \frac{\partial L_{\{AMA\}}}{\partial A}dA$.

bounding both terms yields the stated **Lipschitz-type inequality**.

Let $A \rightarrow A + \Delta A$ and $b \rightarrow b + \Delta b$, where the perturbations satisfy

$$\|\Delta A\|_2 \leq \eta_A, \quad \|\Delta b\|_2 \leq \eta_b.$$

Since $x_{\{AMA\}} = L_{\{AMA\}}b$ in the linear case, the change in $x_{\{AMA\}}$ due to the perturbations is

$$x_{\{AMA\}}(A + \Delta A, b + \Delta b) - x_{\{AMA\}}(A, b) \approx L_{\{AMA\}}\Delta b + \left(\frac{\partial L_{\{AMA\}}}{\partial A}\right)\Delta A.$$

$$\|x_{\{AMA\}}(A + \Delta A, b + \Delta b) - x_{\{AMA\}}(A, b)\|_2 \leq \|L_{\{AMA\}}\|_2 \|\Delta b\|_2 + \|\left(\frac{\partial L_{\{AMA\}}}{\partial A}\right)\|_2 \|\Delta A\|_2..$$

Define a constant $C = \|L_{\{AMA\}}\|_2$.

Then, since $\| \Delta A \|_2 \leq \eta_A, \quad \| \Delta b \|_2 \leq \eta_b$

$$\| x_{\{AMA\}}(A + \Delta A, b + \Delta b) - x_{\{AMA\}}(A, b) \|_2 \leq C(\eta_A + \eta_b).$$

This is the ***Lipschitz-type inequality***: it proves that small perturbations in the inputs (A, b) produce proportionally small changes in the output x$_{AMA}$.

Therefore, **AMA is *stable* and *continuous*** with respect to its input data.☐.

### Deviation from the Moore–Penrose solution

The difference between AMA and the pseudoinverse solution satisfies

$$\| x_{\{AMA\}} - A^+ b \|_2 \leq \| L_{\{AMA\}} - A^+ \|_2 \| b \|_2$$

This positions AMA as an approximate left-inverse operator. The smaller the $\| L_{\{AMA\}} - A^+ \|_2$ the closer AMA behaves to the optimal least-squares estimator.

**Proof** Because $x_{\{AMA\}} = L_{\{AMA\}} b$ and $x_{\{LS\}} = A^+ b \rightarrow x_{\{AMA\}} - x_{\{LS\}} = \left( L_{\{AMA\}} - A^+ \right) b$

and therefore $\| x_{\{AMA\}} - A^+ b \|_2 \leq \| L_{\{AMA\}} - A^+ \|_2 \| b \|_2$
This inequality quantifies precisely how far AMA is from the optimal least-squares operator.

### Non-linear generalization

If AMA is implemented as a smooth non-linear mapping $x_{\{AMA\}}$ = (A, b) with continuous derivative in b, its first-order expansion is:

$$A(A, b + \Delta b) = A(A, b) + J_b A(A, b) \Delta b + O(\| \Delta b \|)_2$$

where J$_b$ (A, b) the Jacobian with respect to b that plays the role of a *local* operator L$_{loc}$.

Setting L$_{loc}$ = J$_b$ (A, b) and replacing L$_{\{AMA\}}$ by L$_{loc}$ in the inequalities above reproduces the same residual and stability bounds locally. Hence, non-linear variants of AMA remain Lipschitz continuous in b and robust to small changes in A (stable).

In summary, the Arithmetic Method Algorithm provides a constructive realization of a stable left-inverse operator for solving linear systems with arbitrary real variables. By replacing explicit matrix inversion with direct arithmetic relations among coefficients and targets, AMA preserves the theoretical properties of least-squares estimation—existence, uniqueness, and bounded residuals—while achieving computational simplicity and robustness under perturbations.

### Connection to implementation

The mathematical framework developed above defines the Arithmetic Method Algorithm (AMA) as a linear (or locally linear) operator L$_{\{AMA\}}$ that approximates the left-inverse of A in the relation Ax = b. The residual, stability, and deviation inequalities establish the theoretical behavior of AMA in terms of accuracy and robustness. In the following section, this theoretical formulation is **numerically implemented** in the form of an **arithmetic algorithm** that reconstructs the target value y through the relations $xj = \frac{y}{ja_{1,j}}$ and $\widehat{y} = \sum_j a_{1,j} x_j$. This implementation realizes the theoretical mapping $x_{\{AMA\}} = L_{\{AMA\}} b$ in component-wise form, enabling direct empirical validation of the theoretical results on stability and residual minimization.

### Numerical validation

To validate the adopted arithmetic method, percentage error is formulated as $\epsilon = \left| \frac{y - \widehat{y}}{y} \right| \cdot 100$, where $y$ is the actual value and $\widehat{y}$ is the computed value by the arithmetic method, while $\epsilon$ is a net number. It holds that when $m \rightarrow 1$, then $\epsilon \rightarrow 0$. Concretely, execution time of computing $\widehat{y}$ is defined as $t \leftarrow t_e - t_s$, where $t_s$ is the starting time and $t_e$ is the ending time, while $t$ is measured in seconds. Intuitively, it also holds that when $m \rightarrow 1$, then $t \rightarrow 0$. An AMA is created to perform numerical validation of the particular solution for $j = 1, \ldots, 1000000$ iterations, which is provided in Table 1. Although, $\epsilon \rightarrow 0$ and $t \rightarrow 0$ for $m \rightarrow 1$, it is observed that $\epsilon$ and $t$ also reach minimum values even when $m \rightarrow 1000000$, which indicates that the proposed arithmetic method has optimal computational behavior.

The Arithmetic Method Algorithm (AMA) described in Tables 2, 3 operationalizes the theoretical operator $L_{\{AMA\}}$ derived previously. While the theoretical formulation treats A and b as general matrix and vector quantities, the numerical version implements their scalar equivalents: a randomly generated coefficient vector $\alpha_{1 \times I}$ represents a row of A, and the scalar y corresponds to b. The algorithm therefore reproduces the linear relation $Ax_{\{AMA\}} \approx b$ in a simplified arithmetic form, allowing numerical evaluation of the residuals and stability properties established in the theoretical section.

AMA takes as an input value $m$, which is the number of polynomial dimensions, while output values of AMA are $I, Y, \widehat{Y}, T, E$ quantities. Specifically, $I$ is the incremental dimension values' vector, $Y$ is the actual values' vector, $\widehat{Y}$ is the predicted values' vector, $T$ is the execution time values' vector, and $E$ percentage error values' vector. To understand the output result quantities of the AMA, $E$ percentage error values' vector and $I$ is the incremental dimension values' vector is visualized in Fig. 1, while $T$ execution time values' vector and $I$ is the incremental dimension values' vector is visualized in Fig. 2. It can be observed that in Fig. 1 maximum

| # | Arithmetic Method Algorithm (AMA) |
|---|---|
| 1 | Input: $m$ // Number of polynomial dimensions, |
| 2 | Output: $I, Y, \hat{Y}, T, E$ // Vectors of $m$ polynomial dimensions, denoting: |
| 3 | // $I$ incremental dimension values' vector, $Y$ actual values' vector, |
| 4 | // $\hat{Y}$ predicted values' vector, $T$ execution time values' vector, and |
| 5 | // $E$ percentage error values' vector |
| 6 | Begin |
| 7 | $\quad m \leftarrow 1{,}000{,}000$ // Set number of polynomial dimensions to iterate |
| 8 | $\quad i \leftarrow 1$ // External loop counter of polynomial dimensions |
| 9 | $\quad$ For $(i \leq m)$ Do |
| 10 | $\quad\quad t_s \leftarrow system.time()$ // Get starting system time |
| 11 | $\quad\quad e \leftarrow 0$ // Set relative error value to 0 |
| 12 | $\quad\quad t \leftarrow 0$ // Set execution time value to 0 |
| 13 | $\quad\quad a_{1 \times i} \leftarrow \emptyset$ // Empty set of polynomial values |
| 14 | $\quad\quad j \leftarrow 1$ // Internal loop counter of polynomial dimensions |
| 15 | $\quad\quad$ While $(j \leq i)$ Repeat |
| 16 | $\quad\quad\quad a_{1,j} \leftarrow random.value(-1000.000, 1000.000)$ // Get a random value withing certain range |
| 17 | $\quad\quad\quad j \leftarrow j + 1$ |
| 18 | $\quad\quad$ End While |
| 19 | $\quad\quad y \leftarrow random.value(-1000.000, 1000.000)$ // Get a random value withing certain range |
| 20 | $\quad\quad x_{1 \times i} \leftarrow \emptyset$ // Empty set of polynomial variables |
| 21 | $\quad\quad j \leftarrow 1$ // Internal loop counter of polynomial dimensions |
| 22 | $\quad\quad$ While $(j \leq i)$ Repeat |
| 23 | $\quad\quad\quad x_{1,j} \leftarrow \frac{y}{j \cdot a_{1,j}}$ |
| 24 | $\quad\quad\quad j \leftarrow j + 1$ |
| 25 | $\quad\quad$ End While |
| 26 | $\quad\quad \hat{y} \leftarrow 0$ // Set predicted value to 0 |
| 27 | $\quad\quad j \leftarrow 1$ // Internal loop counter of polynomial dimensions |
| 28 | $\quad\quad$ While $(j \leq i)$ Repeat |
| 29 | $\quad\quad\quad \hat{y} \leftarrow \hat{y} + (a_{1,j} \cdot x_{1,j})$ |
| 30 | $\quad\quad\quad j \leftarrow j + 1$ |
| 31 | $\quad\quad$ End While |
| 32 | $\quad\quad t_e \leftarrow system.time()$ // Get ending system time |
| 33 | $\quad\quad t \leftarrow t_e - t_s$ // Compute execution time in $i$ polynomial dimension |
| 34 | $\quad\quad \varepsilon \leftarrow \left\lvert \frac{\hat{y} - y}{y} \right\rvert \cdot 100$ // Compute percentage error in $i$ polynomial dimension |
| 35 | $\quad\quad I \leftarrow I \cup i$ // Append current $i$ dimension value to dimension values' vector |
| 36 | $\quad\quad Y \leftarrow Y \cup y$ // Append current $y$ actual value to actual values' vector |
| 37 | $\quad\quad \hat{Y} \leftarrow \hat{Y} \cup \hat{y}$ // Append current $\hat{y}$ predicted value to predicted values' vector |
| 38 | $\quad\quad T \leftarrow T \cup t$ // Append current $t$ execution time value to execution time values' vector |
| 39 | $\quad\quad E \leftarrow E \cup \varepsilon$ // Append current $\varepsilon$ relative error value to relative error values' vector |
| 40 | $\quad$ End For |
| 41 | $\quad Return(I, Y, \hat{Y}, T, E)$ |
| 42 | End |

**Table 1.** Arithmetic Method Algorithm (AMA).

value of $\epsilon$ is 2.1272179572370246e-10, while in Fig. 2 maximum value for $t$ is 4.021273136138916 s, which numerically validates that the results indicate an optimal computational solution even in case where $m$ has an arbitrary maximum value.

(see Appendix A.1 for the correspondence between Theoretical and Algorithmic Variables of the Arithmetic Method Algorithm (AMA)).

### Mathematical foundations of AMR: MSE-optimal combination of AMA and k-NN
Both the theoretical framework and the implemented AMR algorithm are founded on the same principle: a linear combination of the AMA and k–NN estimators, controlled by a blending coefficient α ∈ (0,1). The theoretical part derives the analytically optimal α that minimizes the expected MSE, while the practical implementation estimates or approximates this coefficient empirically through cross-validated performance metrics.

| # | Arithmetic Method Regression (AMR) |
|---|---|
| 1 | Input: $D_{n \times m}$ // Input dataset of n rows and m columns |
| 2 | Output: $MAE_{op}, MSE_{op}, RMSE_{op}, R^2_{op}, alpha_{op}, beta_{op}, delta_{op}, k_{op}, ET$ |
| 3 | // Optimal values of Mean Absolute Error, Mean Squared Error, Root Mean Squared Error, |
| 4 | // R-squared ($R^2$) coefficient, alpha ratio of AMA predicted value, beta ratio of kNN regression |
| 5 | // predicted value, delta for inferring experimentally optimal k nearest neighbors' value, |
| 6 | // k nearest neighbors value, and et execution time |
| 7 | Begin |
| 8 | $MAE_{op} \leftarrow max. value$ // Set a maximum $MAE_{op}$ value |
| 9 | $delta \leftarrow 1$ // Experimentally defined minimum delta value |
| 10 | $delta_{max} \leftarrow 10$ // Experimentally defined maximum delta value |
| 11 | $t_{start} \leftarrow system\ time()$ // Start time |
| 12 | While ($delta \leq delta_{max}$) Repeat // Experimentally defined maximum delta value |
| 13 | $alpha \leftarrow 0.1$ // Experimentally defined minimum alpha AMA regression ratio value |
| 14 | $alpha_{max} \leftarrow 1$ // Experimentally defined maximum alpha AMA regression ratio value |
| 15 | While ($alpha \leq alpha_{max}$) Repeat |
| 16 | $beta \leftarrow 1 - alpha$ // Set beta kNN regression ratio value |
| 17 | $l \leftarrow 1$ //Loop counter for traversing input dataset $D_{n \times m}$ rows |
| 18 | While ($l \leq n$) Repeat // Leave-one-out cross validation evaluation method loop |
| 19 | $X^{tr}_{(n-1) \times (m-1)} \leftarrow D_{n \times m} - D_{1 \times m}$ // Set $X^{tr}$ matrix training set |
| 20 | $Y^{tr}_{(n-1) \times 1} \leftarrow D_{n \times 1} - D_{1 \times 1}$ // Set $Y^{tr}$ matrix training set |
| 21 | $A^{mo}_{(n-1) \times (m-1)}, X^{mo}_{(n-1) \times (m-1)}, Y^{mo}_{(n-1) \times 1} \leftarrow model.function(X^{tr}_{(n-1) \times (m-1)}, Y^{tr}_{(n-1) \times 1})$ |
| 22 | // Get model's $A^{mo}, X^{mo}$ and $Y^{mo}$ matrices |
| 23 | $x^{te}_{1 \times (m-1)} \leftarrow D_{1 \times m}$ // Set $x^{te}$ vector testing set |
| 24 | $y^{ac} \leftarrow D_{1 \times m}$ // Set current $y^{ac}$ actual value |
| 25 | $\hat{y}^{pr}, k \leftarrow predict.function(A^{mo}_{(n-1) \times (m-1)}, X^{mo}_{(n-1) \times (m-1)}, Y^{mo}_{(n-1) \times 1}, x^{te}_{1 \times (m-1)}, alpa, beta, delta)$ |
| 26 | // Get $\hat{y}^{pr}$ current predicted value as a combination of AMA and kNN predicted values, and |
| 27 | // k number of neighbors used for current prediction |
| 28 | $Y^{ac} \leftarrow Y^{ac} \cup y^{ac}$ // Append current $y^{ac}$ actual value to actual values' vector |
| 29 | $\hat{Y}^{pr} \leftarrow \hat{Y}^{pr} \cup \hat{y}^{pr}$ // Append current $\hat{y}^{pr}$ predicted value to predicted values' vector |
| 30 | $l \leftarrow l + 1$ |
| 31 | End While |
| 32 | $MAE \leftarrow MAE.function(Y^{ac}, \hat{Y}^{pr})$ // Compute current MAE, invoking Python Library Code |
| 33 | $MSE \leftarrow MSE.function(Y^{ac}, \hat{Y}^{pr})$ // Compute current MSE, invoking Python Library Code |
| 34 | $RMSE \leftarrow RMSE.function(Y^{ac}, \hat{Y}^{pr})$ //Compute current RMSE, invoking Python Library Code |
| 35 | $R^2 \leftarrow R^2.function(Y^{ac}, \hat{Y}^{pr})$ // Compute current $R^2$, invoking Python Library Code |
| 36 | If ($MAE \leq MAE_{op}$) Then // Check for optimal evaluation metrics' values |
| 37 | $MAE_{op} \leftarrow MAE$ // Get optimal MAE |
| 38 | $MSE_{op} \leftarrow MSE$ // Get optimal MSE |
| 39 | $RMSE_{op} \leftarrow RMSE$ // Get optimal RMSE |
| 40 | $R^2_{op} \leftarrow R^2$ // Get optimal $R^2$ |
| 41 | $alpha_{op} \leftarrow alpha$ // Get optimal alpha |
| 42 | $beta_{op} \leftarrow beta$ // Get optimal beta |
| 43 | $delta_{op} \leftarrow delta$ // Get optimal delta |
| 44 | $k_{op} \leftarrow k$ // Get optimal k |
| 45 | End If |
| 46 | $alpha \leftarrow alpha + 0.1$ |
| 47 | End While |
| 48 | $delta \leftarrow delta + 0.1$ |
| 49 | $t_{end} \leftarrow system\ time()$ // End time |
| 50 | $ET \leftarrow t_{end} - t_{start}$ // Compute execution time |
| 51 | End While |
| 52 | Return($MAE_{op}, MSE_{op}, RMSE_{op}, R^2_{op}, alpha_{op}, beta_{op}, delta_{op}, k_{op}, ET$) |
| 53 | End |

**Table 2.** Arithmetic method regression (AMR).

AMR combines AMA and k-NN predictions as

$$\hat{y}_{AMR}(x) = \alpha * \hat{y}_{AMA}(x) + (1 - \alpha) * \hat{y}_{kNN}(x)$$

Define $U = \hat{y}_{AMA}(x)$, $V = \hat{y}_{kNN}(x)$, and the expected MSE risk

```
#    Model Function Algorithm (MFA)
1    Input: X^tr_(n-1)×(m-1), Y^tr_(n-1)×1  // Input training regressors' and regressand matrices
2    Output: A^mo_(n-1)×(m-1), X^mo_(n-1)×(m-1), Y^mo_(n-1)×1
3         // Output model polynomial values, regressors' and regressand matrices
4    Begin
5        l ← 1  // Loop counter for traversing input training sets n − 1 rows
6        While (l ≤ (n − 1)) Repeat  // Loop ends when l traverses all the rows
7            x_1×(m-1) ← X^tr_l×(m-1)  // Initialize current training regressors' vector from regressors' matrix
8            y ← Y^tr_l×1  // Initialize current training regressand value from regressand matrix
9            a_1×(m-1), x_1×(m-1), ŷ ← core.function(x_1×(m-1), y)  // Invoke a customized AMA function
10           A^mo_l×(m-1) ← A^mo_l×(m-1) ∪ a_1×(m-1)  // Model matrix of computed polynomial values based on AMA
11           X^mo_l×(m-1) ← X^mo_l×(m-1) ∪ x_1×(m-1)  // Model matrix of regressors' values based on AMA
12           Y^mo_l×1 ← Y^mo_l×1 ∪ ŷ  // Model matrix of computed regressand predicted values based on AMA
13       End While
14       Return(A^mo_(n-1)×(m-1), X^mo_(n-1)×(m-1), Y^mo_(n-1)×1)
15   End
```

**Table 3.** Model function algorithm (MFA).

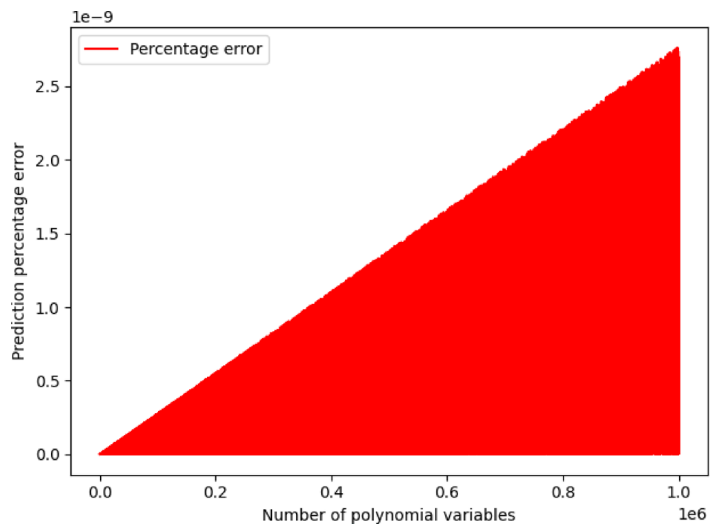

**Fig. 1.** Visualization of AMA percentage error values for 1,000,000 iterations.

$$R(\alpha) = E\left[(\alpha U + (1-\alpha)V - Y)^2\right]$$

While the theoretical derivation assumes a quadratic loss (MSE), the algorithm employs MAE as a practical and robust empirical proxy for model error minimization. Since both MAE and MSE are convex loss functions that measure average prediction deviation, minimizing MAE in practice leads to similar optimal blending behavior as minimizing MSE, with the added advantage of reduced sensitivity to outliers.

The iterative AMR procedure, which loops over α and δ while evaluating prediction error, constitutes a discrete empirical minimization of the expected risk R(α). This iterative search operationalizes the theoretical minimization of R(α) by replacing the analytic derivative condition (∂R/∂α = 0), that will give the optimal solution α*, with a numerical optimization over a finite grid of α values.

Theoretically, if $E\left[(U-V)^2\right] > 0$ the unique minimizer of R(α) is:

$$\alpha^* = E\frac{[(Y-V)(U-V)]}{E[(U-V)^2]} = \frac{Cov(U-V, Y-V)}{Var(U-V)}$$

(see Appendix A.2 for the full derivation of α*)

In finite samples, the expectations in the expression for α* are unknown. They can be estimated empirically through the unbiased sample estimator:

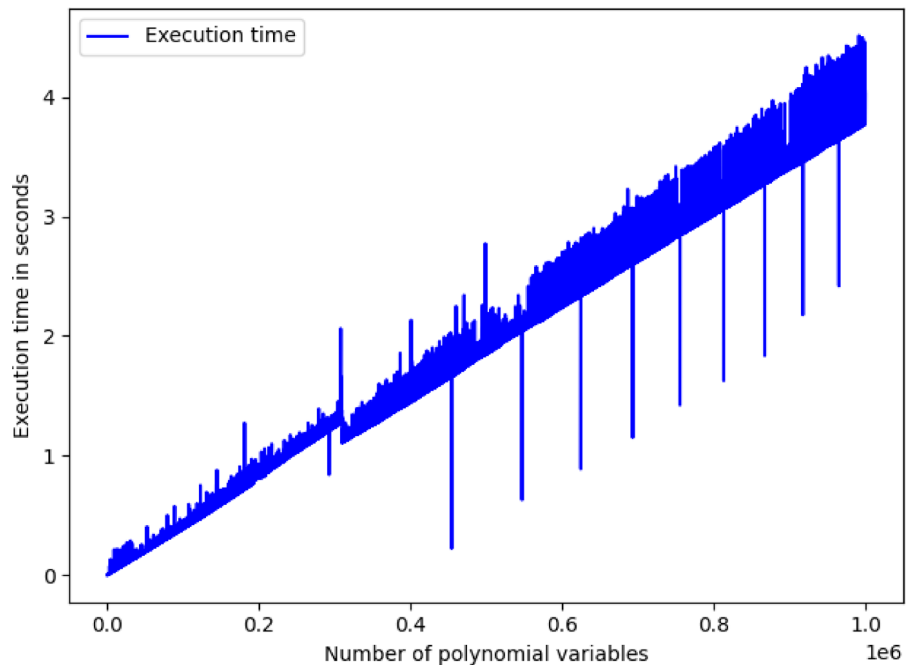

**Fig. 2.** Visualization of AMA execution time values for 1,000,000 iterations.

$$\widehat{\alpha} = \frac{\Sigma_i (y_i - v_i)(u_i - v_i)}{\Sigma_i (u_i - v_i)^2}$$

In the implemented AMR algorithm, this estimator is approximated by a discrete grid search over α values that minimizes the empirical error (MAE). This grid-search procedure can be interpreted as a discrete, empirical approximation to the theoretical minimization of the expected MSE. The use of MAE is considered as a **robust empirical adaptation** of the theoretical MSE objective.

In summary, both the theoretical and empirical formulations of AMR pursue the same goal: identifying the blending coefficient α that minimizes prediction error. The theoretical estimator $\widehat{\alpha}$ computes this coefficient directly from covariance and variance terms, whereas the algorithm approximates it through a discrete empirical search minimizing MAE. Consequently, $α_{op}$ serves as a numerical approximation to $\widehat{\alpha}$, yielding equivalent behavior when the grid resolution is sufficiently fine.

### Bias–variance tradeoff and consistency of AMR

Having established the theoretical and empirical mechanisms for determining the optimal blending weight α, we now analyze how this combination improves predictive performance through the classical bias–variance framework.

The generalization performance of AMR can be interpreted through the classical bias–variance decomposition of the mean-squared error (MSE):

$$E\left[\left(\widehat{f}(x) - y\right)^2\right] = \left(E\left(\widehat{f}(x) - f(x)\right)^2 + Var\left(\widehat{f}(x)\right) + \sigma^2 \; \forall \; \text{estimator} \; \widehat{f}(x)\right.$$

where $f(x) = E[Y \mid X = x]$ the true regression function and $σ^2$ the irreducible noise.

The Arithmetic Method Algorithm (AMA) produces predictions by direct algebraic operations on the data matrix A and the vector b. Because this mapping is deterministic and stable under small perturbations of the input, AMA has *low estimator variance* (its output changes minimally across different samples drawn from the same distribution). However, due to its algebraic and mostly linear structure, AMA may not fully capture nonlinear dependencies in the data, resulting in *moderate bias* with respect to the true f(x).

In contrast, the k–nearest neighbors (k–NN) estimator is highly flexible and data-driven. It can approximate nonlinear functions arbitrarily well when sufficient data are available, implying *low bias*. Yet, because the local averaging depends directly on random neighborhood composition, it tends to have *high variance*.

The **Arithmetic Method Regression (AMR)** model combines these two complementary estimators $\widehat{y}_{\{AMA\}}$ and $\widehat{y}_{\{kNN\}}$ through a *convex blending weight* α. The optimal coefficient $α^*$ minimizes the expected MSE R(α), effectively balancing the bias–variance tradeoff between AMA and k–NN. For finite samples, this value is estimated empirically by $\widehat{\alpha}$ demonstrating that AMR achieves a *lower overall error than either component alone*, because α* shifts weight toward the lower-variance AMA component in small samples and toward the lower-bias k–NN component as sample size increases.

In this theoretical model, the term $\widehat{y}_{kNN}(x)$ denotes the k–NN regression prediction. In practice, the δ parameter in the AMR algorithm modifies the neighborhood definition used by k–NN, effectively controlling the smoothness and variance of V. Thus, δ indirectly influences the optimal value of α by changing the variance and bias properties of the k–NN component. Hence, δ can be viewed as a practical extension that fine-tunes the empirical estimate of α* by calibrating the bias–variance tradeoff in V. Empirically, δ plays a complementary role to α, ensuring that the bias–variance equilibrium derived theoretically is preserved under practical neighborhood tuning.

From a theoretical perspective, consistency of AMR follows from classical results for k–NN regression[24–26]: if k→∞ and k/n→0 as n→∞, the k–NN estimator converges in probability to f(x). Since AMA defines a continuous and Lipschitz mapping in b, the convex combination of a consistent estimator (k–NN) and a continuous estimator (AMA) remains consistent under these same asymptotic conditions. Hence, AMR inherits both finite-sample bias–variance efficiency and asymptotic consistency.

In summary, the theoretical derivation of α* establishes the MSE-optimal blending rule between AMA and k–NN, while the empirical AMR algorithm implements a discrete, MAE-based approximation of the same principle. The grid-search optimization over α and δ can thus be viewed as a practical realization of the theoretical minimization of R(α), ensuring that the empirical $α_{op}$ aligns closely with the theoretically optimal â derived from MSE analysis. Together, these derivations establish the full mathematical validity and practical continuity of the AMR framework, linking theoretical MSE-optimality with empirically verified performance. Overall, this theoretical–empirical alignment ensures that the AMR algorithm's practical optimization procedure is a faithful numerical realization of the analytically MSE-optimal combination, guaranteeing both mathematical soundness and empirical robustness.

### Arithmetic method regression (AMR) algorithm

To optimize k-Nearest Neighbors (k-NN) for regression problems an AMR algorithm is proposed, which incorporates the adopted arithmetic method introduced in the current research effort for performing linear regression. Specifically, AMR predicts a computed $\widehat{y}^{pr}$ numerical value, where the predicted value is based on a ratio defined between AMA used for linear regression and k-NN regression algorithms' values, respectively. Concretely, $alpha$ value indicates the contribution of the computed $\widehat{y}^{pr}_{AMA}$ value inferred by AMA, while $beta$ value indicates the contribution of the computed $\widehat{y}^{pr}_{kNN}$ value inferred by k-NN. Relation between $alpha$ value and $beta$ value is set according to the equation: $alpha = 1 - beta$ as part of the AMR iterative process to assess the efficiency of the partially contributed AMA and k-NNN regression algorithms. Subsequently, $alpha$ and $beta$ take values within the intervals $alpha \in (0, 1]$ and $beta \in [0, 1)$, where adopted values are experimentally increased with step of 0.1. It holds that the predicted regressand value of AMR is provided by the equation: $\widehat{y}^{pr} \leftarrow (alpha \cdot \widehat{y}^{pr}_{AMA}) + (beta \cdot \widehat{y}^{pr}_{kNN})$, as discussed in Predict Function Algorithm (PFA) presented consecutively. Intuitively, AMR incorporates a $delta$ heuristic numerical value, which is used to allow slightly farther $k$ neighbors to be considered close enough during the prediction process thus calibrating heuristically the optimal number of $k$ neighbors adopted by k-NN regression algorithm. Proposed AMR is presented in Table 4, where input value is a $D_{n \times m}$ input dataset of $n$ rows and $m$ columns, while output results are optimal values used to assess the efficiency of the proposed AMR. Specifically, $MAE_{op}$ indicates an optimal value for Mean Absolute Error, $MSE_{op}$ indicates an optimal value for Mean Squared Error, $RMSE_{op}$ indicates an optimal value for Root Mean Squared Error, $R^2{}_{op}$ indicates and optimal value for R-squared ($R^2$) coefficient, $alpha_{op}$ indicates an optimal value for $alpha$ AMA contribution in prediction value, $beta_{op}$ indicates an optimal value for $beta$ k-NN contribution is prediction value, $delta_{op}$ indicates an optimal value of $delta$ for inferring heuristically optimal $k$ nearest neighbors' value, while $k_{op}$ indicates an optimal value for $k$ nearest neighbors value, and $ET$ is the Execution Time required to run the algorithm and get the output results. Concretely, evaluation metric to infer AMR optimal results is relevant to the value of $MAE_{op}$ which is initially set to the maximum numerical value provided by the execution programming environment. It holds that the adopted evaluation metric of the proposed AMR algorithm is expressed as a relation between the observed $MAE$ of a certain algorithm iteration compared with the current $MAE_{op}$. Continually, at each algorithm iteration it is performed a comparison between the observed $MAE$ value and the current $MAE_{op}$ value. The value of current $MAE_{op}$ is updated to that of the observed $MAE$ value only if the observed $MAE$ of a certain iteration is less than or equal to the current value of $MAE_{op}$. In this iterative process, the adopted optimization values are updated accordingly and output the AMR optimal results. It should be noted that $MAE$ is adopted as a cost evaluation metric that needs to be minimized, since it is not as heavily influenced by outliers as $MSE$[27]. Additionally, MAE is also adopted as a cost evaluation metric for accessing performance in the current research effort, since it is dealing efficiently when experimented with k-NN where in this context is the basis algorithm that is optimized by the proposed AMR algorithm[28].

| # | Core Function Algorithm (CFA) |
|---|---|
| 1 | Input: $x_{1 \times (m-1)}, y$ // Input training regressors' vector and regressand value based on AMA function |
| 2 | Output: $a_{1 \times (m-1)}, x_{1 \times (m-1)}, \widehat{y}$ // Output computed polynomial values' vector, regressors' vector, and |
| 3 | // regressand predicted value |
| 4 | Begin |
| 5 | $i \leftarrow 1$ //Loop counter for traversing input training vectors' dimensions (i.e., $m - 1$ columns) |
| 6 | While ($i \leq (m - 1)$) Repeat // Loop ends when $i$ traverses all the columns |
| 7 | $a_{1,i} \leftarrow \frac{y}{i \cdot x_{1,i}}$ // Compute current polynomial value from regressors' and regressand matrices |
| 8 | $a_{1 \times (m-1)} \leftarrow a_{1 \times (m-1)} \cup a_{1,i}$ //Append current $a_{1,i}$ polynomial value to polynomial values' vector |
| 9 | $x_{1 \times (m-1)} \leftarrow x_{1 \times (m-1)} \cup x_{1,i}$ //Append current $x_{1,i}$ regressor value to regressors' vector |
| 10 | $\widehat{y} \leftarrow \widehat{y} + (a_{1,i} \cdot x_{1,i})$ // Compute regressand predicted value |
| 11 | End While |
| 12 | Return $(a_{1 \times (m-1)}, x_{1 \times (m-1)}, \widehat{y})$ |
| 13 | End |

**Table 4.** Core function algorithm (CFA).

*(see Appendix A.3 for the correspondence between Theoretical and Algorithmic Variables of the Algorithmic Method Regression (AMR)).*

It can be observed thar AMR invokes certain functional algorithms implemented in this research effort, such as Model Function Algorithm (MFA) used to build the proposed linear prediction model by invoking Core Function Algorithm (CFA), which is a customized AMA function adopted in the proposed approach. AMR also incorporates Predict Function Algorithm (PFA) necessary for performing linear regression and infer $\widehat{y}^{pr}$ value. It can be also observed that AMR invokes programming environment's available functions for defining optimal $mae$, $mse$, $rmse$, and $r2$ values. Specifically, MFA takes as input the values of training regressors' $X^{tr}_{(n-1)\times(m-1)}$ and regressand $Y^{tr}_{(n-1)\times 1}$ matrices, which are extracted from the $D_{n\times m}$ input dataset of $n$ rows and $m$ columns, respectively. Concretely, the output values of MFA are the values which compose the result model that contains the polynomial values $A^{mo}_{(n-1)\times(m-1)}$, regressors' $X^{mo}_{(n-1)\times(m-1)}$ and regressand $Y^{mo}_{(n-1)\times 1}$ matrices, respectively. Proposed MFA is presented in Table 3. Subsequently, CFA takes as input the $x_{1\times(m-1)}$ and $y$ values, which are the input training regressors' vector and regressand value based on AMA function. The output values of CFA are the resulted computed polynomial values' vector $a_{1\times(m-1)}$, the regressors' vector $x_{1\times(m-1)}$ and the regressand predicted value $\widehat{y}$, respectively. Adopted CFA is presented in Table 4. Continually, PFA performs the proposed linear regression prediction of the current research effort. Specifically, input values of PFA are the input model polynomial values' matrix $A^{mo}_{(n-1)\times(m-1)}$, the regressors' model matrix $X^{mo}_{(n-1)\times(m-1)}$, the regressand model vector $Y^{mo}_{(n-1)\times 1}$, the regressor test vector $x^{te}_{1\times(m-1)}$, the current $alpha$ value, the current $beta$ value and the current $delta$ value. Subsequently, the output result values of PFA are the output computed predicted regressand value $\widehat{y}^{pr}$ and the $k$ number of neighbors invoked for current regression output. Intuitively, to define the optimal $k$ it is used the following conditional inequality: $dist_k < (delta \cdot dist_{min})$, where $dist_{min}$ finds the closest single neighbor in the given dataset by multiplying $dist_{min}$ with $delta$ to allow neighbors that are slightly further than the absolute closest to also be counted. It holds that this conditional inequality selects heuristically the optimal neighbors with regards to all training points that are within a threshold distance of the nearest training point. Continuously, $delta$ takes values within the interval $delta \in [1, 10]$, where adopted values are experimentally increased with step of $0.1$. Concretely, due to computational complexity limitations, which are discussed on the Results section, adopted experimental real world data sources are rather small. Subsequently, due to the small dataset sizes prediction process is prone to outliers, thus is adopted the Manhattan (i.e., City block) distance for computing the distances $dist_k$ and $dist_{min}$, since it is more robust in presence of outliers[29]. Proposed PFA is presented in Table 5.

## Results
### Data sources
Efficiency of the proposed AMR algorithm is assessed by applying it to real world data sources. Specifically, there are 17 separate data sources used, where certain preprocessing is applied to transform all provided data attributes to numeric regressors' and regressand values, necessary to input proposed AMR algorithm. From the 17 total data sources the first 12 are small, where their length is within the interval of $[43, 297]$ instances. Intuitively, to be able to generalize the results to bigger datasets we incorporated 5 other data sources, where their length is within the interval of $[398, 1178]$ instances. Additionally, instances with missing values were removed from the initial datasets. Consequently, a dimensionality reduction algorithm is applied, which evaluates the worth of a subset of regressors' values by considering the individual predictive ability of each regressor's value along with the degree of redundancy between them. Subsets of regressors' values that are highly correlated with the predicted regressand value while having low intercorrelation are preferred for further experimentation[30]. Intuitively, by applying such a dimensionality reduction process is leading to more meaningful data sources, while the selected regressors are a subset of the initial set of regressors provided by the preprocessed data sources. Concretely, the first data source, i.e., Data-1, concerns the study of the factors affecting patterns of insulin-dependent diabetes mellitus in children. The objective is to investigate the dependence of the level of serum C-peptide on the various other factors to understand patterns of residual insulin secretion. The response measurement is the logarithm of C-peptide concentration (pmol/ml) at the diagnosis, and the predictor measurements age and base deficit, a measure of acidity. Regressors have numerical values in total. There were no instances with missing values in the initial dataset. Data source size is 43 instances, where number of selected regressor values is 2, along with a regressand value used for prediction. Subsequently, the second data source, i.e., Data-2, consists of learning quantitative structure activity relationships (QSARs) for defining the inhibition of dihydrofolate reductase by pyrimidines. Intuitively, data source describes a drug design by incorporating data analytics for the use of inductive logic programming to model the structure-activity relationships of trimethoprim analogues binding to dihydrofolate reductase. Regressors have numerical values in total. There were no instances with missing values in the initial dataset. Data source size is 74 instances, where number of selected regressor values is 6, along with a regressand value used for prediction. Additionally, the third data source, i.e., Data-3, contains data where each instance represents follow-up data for one breast cancer case. These are initialized by consecutive patients and include only those cases exhibiting invasive breast cancer and no evidence of distant metastases at the time of diagnosis. Certain regressors' values are computed from a digitized image of a fine needle aspirate (FNA) of a breast mass. They describe characteristics of the cell nuclei present in the image. Regressors have numerical values in total. There were no instances with missing values in the initial dataset. Data source size is 194 instances, where number of selected regressor values is 16, along with a regressand value used for prediction. Continuously, the fourth data source, i.e., Data-4, provides estimations of the percentage of body fat determined by underwater weighing and various body circumference measurements for certain men participants. Such dataset can be used to apply multiple regression techniques to provide an accurate measurement of body fat, which should be a convenient and desirable prediction behavior compared with other methods that do not result in optimal output. Regressors have numerical values in total. There were no instances with missing values in the initial dataset. Data source size is 252 instances, where number of selected regressor values is 3, along with a regressand value used for prediction.

| # | Predict Function Algorithm (PFA) |
|---|---|
| 1 | Input: $A^{mo}_{(n-1)\times(m-1)}, X^{mo}_{(n-1)\times(m-1)}, Y^{mo}_{(n-1)\times 1}, x^{te}_{1\times(m-1)}$, alpa, beta, delta |
| 2 | // Input model polynomial values' matrix, regressors' model matrix, |
| 3 | // regressand model vector, regressor test vector, and alpa, beta, delta current values |
| 4 | Output: $\hat{y}^{pr}, k$ // Output computed predicted regressand value, and |
| 5 | // number of neighbors invoked for current regression output |
| 6 | Begin |
| 7 | $k \leftarrow 0$ // Set initial neighbors value to zero |
| 8 | $dist_{min} \leftarrow max.value$ |
| 9 | // Set a maximum $dist_{min}$ value (i.e. supportive distance used for optimizing neighbors' distance) |
| 10 | $l \leftarrow 1$ // Loop counter for traversing input models' matrices rows (i.e., $n-1$ rows) |
| 11 | For $(l \leq (n-1))$ Do |
| 12 | $\quad dist_{par} \leftarrow 0$ // Set neighbors partial distance to zero |
| 13 | $\quad i \leftarrow 1$ //Loop counter for traversing input models' matrices dimensions (i.e., $m-1$ columns) |
| 14 | $\quad$ While $(i \leq (m-1))$ Repeat // Loop ends when $i$ traverses all the columns |
| 15 | $\quad\quad dist_{par} \leftarrow dist_{par} + \left| x^{te}_{1,i} - X^{mo}_{l,i} \right|$ // Compute current partial neighbors' distance |
| 16 | $\quad\quad i \leftarrow i+1$ |
| 17 | $\quad$ End While |
| 18 | $\quad$ If $(dist_{par} < dist_{min})$ Then |
| 19 | $\quad\quad dist_{min} \leftarrow dist_{par}$ // Update supportive neighbors' distance |
| 20 | $\quad$ End If |
| 21 | End For |
| 22 | $l \leftarrow 1$ // Loop counter for traversing input models' matrices rows (i.e., $n-1$ rows) |
| 23 | For $(l \leq (n-1))$ Do |
| 24 | $\quad dist_k \leftarrow 0$ // Set optimal neighbors' distance to zero |
| 25 | $\quad i \leftarrow 1$ //Loop counter for traversing input models' matrices dimensions (i.e., $m-1$ columns) |
| 26 | $\quad$ While $(i \leq (m-1))$ Repeat // Loop ends when $i$ traverses all the columns |
| 27 | $\quad\quad dist_k \leftarrow dist_k + \left| x^{te}_{1,i} - X^{mo}_{l,i} \right|$ // Compute current optimal neighbors' distance |
| 28 | $\quad\quad i \leftarrow i+1$ |
| 29 | $\quad$ End While |
| 30 | $\quad$ If $(dist_k < (delta \cdot dist_{min}))$ Then // Heuristically infer the optimal neighbors' distance |
| 31 | $\quad\quad \hat{y}^{cu}_{AMA} \leftarrow 0$ // Set proposed current AMA predicted value to zero |
| 32 | $\quad\quad \hat{y}^{cu}_{kNN} \leftarrow 0$ // Set current kNN regression predicted value to zero |
| 33 | $\quad\quad q \leftarrow l$ // Set index (i.e., $q$) of optimal neighbor in current row $l$ |
| 34 | $\quad\quad k \leftarrow k+1$ // Update optimal number of neighbors |
| 35 | $\quad\quad i \leftarrow 1$ //Loop counter for traversing input models' matrices dimensions (i.e., $m-1$ columns) |
| 36 | $\quad\quad$ While $(i \leq (m-1))$ Repeat // Loop ends when $i$ traverses all the columns |
| 37 | $\quad\quad\quad \hat{y}^{cu}_{AMR} \leftarrow \hat{y}^{cu}_{AMR} + \left( A^{mo}_{q,i} \cdot x^{te}_{1,i} \right)$ |
| 38 | $\quad\quad\quad$ // Compute current AMA predicted value (i.e., based on customized AMA algorithm) |
| 39 | $\quad\quad$ End While |
| 40 | $\quad\quad \hat{y}^{cu}_{kNN} \leftarrow \hat{y}^{cu}_{kNN} + Y^{mo}_{q,1}$ // Compute current kNN regression predicted value |
| 41 | $\quad$ End If |
| 42 | $\quad \hat{y}^{pr}_{AMA} \leftarrow \hat{y}^{pr}_{AMA} + \hat{y}^{cu}_{AMA}$ // Add current AMA predicted value to optimal AMA predicted value |
| 43 | $\quad \hat{y}^{pr}_{kNN} \leftarrow \hat{y}^{pr}_{kNN} + \hat{y}^{cu}_{kNN}$ // Add current kNN predicted value to optimal kNN predicted value |
| 44 | End For |
| 45 | $\hat{y}^{pr} \leftarrow \left( alpha \cdot \hat{y}^{pr}_{AMA} \right) + \left( beta \cdot \hat{y}^{pr}_{kNN} \right)$ // Computed AMR predicted regressand value |
| 46 | Return$(\hat{y}^{pr}, k)$ |
| 47 | End |

**Table 5.** Predict function algorithm (PFA).

Intuitively, the fifth data source, i.e., Data-5, contains tumor sizes of breast cancer that have repeatedly appeared in the machine learning literature treating cases of lymphography and primary-tumor incidents. Adopted data sources contain instances of negative recurrence breast tumor distribution events as well as instances of positive recurrence breast tumor distribution events. Regressors are both numerical and nominal, where preprocessing transformed the nominal regressors into numerical values. Instances with missing values were removed from the initial dataset. Data source size is 277 instances, where number of selected regressor values is 6, along with a regressand value used for prediction. Subsequently, the sixth data source, i.e., Data-6, describes the contents of a medical heart-disease directory, where focus is given on predicting cholesterol levels in provided patients' blood tests. Regressors are both numerical and nominal, where preprocessing transformed the nominal regressors into numerical values. Instances with missing values were removed from the initial dataset. Data source size is 297 instances, where number of selected regressor values is 6, along with a regressand value used for prediction. Continuously, the seventh data source i.e., Data-7, contains patients who suffered heart attacks at some point in the past, where some are still alive and some are not. Specifically, the survival and still-alive regressors' values indicate whether a patient survived for at least one year following the heart attack. Challenge focuses on the prediction of whether or not the patient will survive at least one year, while the most difficult part of this regression problem is to efficiently predict that the patient will not survive. Regressors are both numerical and nominal, where preprocessing transformed the nominal regressors into numerical values. Instances with missing values were removed from the initial dataset. Data source size is 61 instances, where number of selected regressor values is 3, along with a regressand value used for prediction. Concretely, the eighth data source, i.e., Data-8, focuses on the identification of risk factors associated with giving birth to a low-birth-weight baby (weighing less than 2500 g), where data were collected on a certain population of women. Intuitively, some of them had low birth weight babies, while the others had normal birth weight babies. Specifically, low birth weight is an outcome that has been of concern to physicians in recent years. This is explained since infant mortality rates and birth defect rates are very high for low-birth-weight babies. Additionally, a woman's behavior during pregnancy (including diet, smoking habits, and receiving prenatal care) can have a significant impact on the chances of carrying the baby to terms and, consequently, of delivering a baby of normal birth weight. Regressors are both numerical and nominal, where preprocessing transformed the nominal regressors into numerical values. There were no instances with missing values in the initial dataset. Data source size is 189 instances, where number of selected regressor values is 4, along with a regressand value used for prediction.

Subsequently, the nineth data source, i.e., Data-9, contains data recorded from the Mayo Clinic trial in primary biliary cirrhosis (PBC) of the liver conducted between 1974 and 1984. Specifically, between January 1974 and May 1984, the Mayo Clinic conducted a double-blinded randomized trial in primary biliary cirrhosis of the liver (PBC), comparing the drug D-penicillamine (DPCA) with a placebo on certain patients. The data from the trial were analyzed in 1986 for presentation in clinical literature. For that analysis, disease and survival status as of July 1986, were recorded for as many patients as possible. Regressors are both numerical and nominal, where preprocessing transformed the nominal regressors into numerical values. Instances with missing values were removed from the initial dataset. Data source size is 276 instances, where number of selected regressor values is 5, along with a regressand value used for prediction. Additionally, the tenth data source, i.e., Data-10, provides data for a part of a large clinical trial carried out by the Radiation Therapy Oncology Group in the United States. Specifically, full study included patients with squamous carcinoma in the mouth and throat. The patients entered the study were randomly assigned to one of two treatment groups, radiation therapy alone or radiation therapy together with a chemotherapeutic agent. The objective of the study was to compare the two treatment policies with respect to patient survival. Regressors are both numerical and nominal, where preprocessing transformed the nominal regressors into numerical values. Instances with missing values were removed from the initial dataset. Data source size is 193 instances, where number of selected regressor values is 4, along with a regressand value used for prediction. Concretely, the eleventh data source, i.e., Data-11, contains several pollution regressors' variables that are extensively used in the literature for performing regression analysis to investigate the relation between air pollution and human mortality. Regressors have numerical values in total. There were no instances with missing values in the initial dataset. Data source size is 60 instances, where number of selected regressor values is 5, along with a regressand value used for prediction. Intuitively, the twelfth data source, i.e., Data-12, provides data collected by patients participated in an administration of Lung Cancer Trial (LCT). Concretely, the aim is to predict the patients' survival time. Regressors have numerical values in total. There were no instances with missing values in the initial dataset. Data source size is 137 instances, where number of selected regressor values is 4, along with a regressand value used for prediction.

Consequently, the thirteenth data source, i.e., Data-13, provides data relevant to abalone. The aim is to predict the age of abalone from physical measurements. Supporting data measurements incorporate operations of determining the age of abalone regarding the cut of the shell through the cone, straining it, and counting the number of rings through a microscope. Regressors are both numerical and nominal, where preprocessing transformed the nominal regressors into numerical values. There were no instances with missing values in the initial data dataset. Data source size is 1063 instances, where number of selected regressor values is 2, along with a regressand value used for prediction. Subsequently, the fourteenth data source, i.e., Data-14, provides data that concerns the city cycle fuel consumption of a vehicle measured in travel miles per gallon of oil. Focus is given on predicting the value of miles per gallon for certain types of vehicles, which contain adequate data for describing the profile of each participating vehicle. Regressors are both numeric and nominal, where preprocessing transformed nominal to regressors into numerical values. There were 6 instances with missing values, which are omitted from the initial dataset. Data source size is 398 instances, where number of selected regressor values is 7 along with a regressand value used for prediction. Concretely, the fifteenth data source, i.e., Data-15, provides data relevant to daily stock prices in a certain period between January 1988 through October 1991 for 10 aerospace companies. Focus is given on predicting companies' stock prices based on prior knowledge implicit is the data source. Regressors have numerical values in total. There were no instances with missing values in the initial dataset. Data source size is 754 instances, where the number of the selected regressor values is 2 along with the regressand value used for prediction. Intuitively, the sixteenth data source, i.e., Data-16, provides data that contains seismic event records characterized by focal depth, geographic coordinates (i.e., latitude and longitude), and earthquake magnitude on the Richter scale. Focus is given on predicting earthquake magnitude in relation to focal depth and spatial location. Regressors have numerical values in total. There were no instances with missing values in the initial dataset. Data source size is 1178 instances, where the number of the selected regressor values is 1 along with the regressand value used for prediction. Additionally, the seventeenth data source, i.e., Data-17, provides data that consists of annual observations on the level of strike volume in days and their covariates in the Organization for Economic Co-operation and Development (OECD) countries in the period between 1951 and 1985. Focus is given on predicting the strike volume based on stochastic data. Regressors are both numerical and nominal, where preprocessing transformed the nominal regressors into numerical values. There were no instances with missing values in the initial data dataset. Data source size is 625 instances, where the number of the selected regressor values is 2 along with the regressand value used for prediction. All data sources evaluated are publicly available in the data availability section.

### Evaluation

Since adopted data sources are small to evaluate the proposed AMR algorithm it is incorporated the leave-one-out cross validation evaluation method. Such method is used mainly when the available data sources are relatively small[31]. During the leave-one-out cross validation a data source is divided into subsets, where each subset contains only one instance. Concretely, in an iteration loop, which repeats for every single subset al.l subsets except of one (i.e., at each time) are used as a training set while the remaining single subset is used as a testing set. It holds that at the end of the iteration process all subsets have been used both as part of a training set as well as a testing set, respectively. Intuitively, such evaluation method is applied to all of the data sources provided while at the end of the process the results are the predicted numerical values assigned to the regressand variables. Given the predicted and the actual values of the regressand variables of each data source, certain evaluation metrics are applied to assess the efficiency of the proposed algorithm. Such evaluation metrics are MAE, MSE, RMSE, and $R^2$. Specifically, MAE is a common evaluation metric for assessing the efficiency of regression algorithms and defined as follows. Given a set of predicted regressand values, $\widehat{y_l}$, and the actual regressand values, $y_l$, for a dataset of size $n$ rows (i.e., $l = 1, \ldots, n$), MAE is computed as: $MAE = \frac{1}{n} \sum_{l=1}^{n} |y_l - \widehat{y_l}|$, which measures the average magnitude of errors between predicted and actual values, where errors are considered in absolute value while units are the same as the predicted regressand variable. MSE is another evaluation metric for regression analysis, which is defined as follows. Given a set of predicted regressand values, $\widehat{y_l}$, and the actual regressand values, $y_l$, for a data source of size $n$ rows (i.e., $l = 1, \ldots, n$), MSE is computed as: $MSE = \frac{1}{n} \sum_{l=1}^{n} (y_l - \widehat{y_l})^2$, which measures the average squared difference between predicted values and actual values. Concretely, squaring converts all errors to positive values and penalizes large errors more strongly than the small ones, thus the units are the squared of the predicted regressand variable. RMSE is also an evaluation metric used in regression problems, which is defined as follows. Given a certain MSE the RMSE is the square root of the given MSE and is computed as: $RMSE = \sqrt{\frac{1}{n} \sum_{l=1}^{n} (y_l - \widehat{y_l})^2}$, which measures the average size of the predicted regressand errors providing focus on large errors because of the squaring. In addition, units are the same as the original variable, which provides a more interpretable error explanation than the MSE. It holds that a lower RMSE denotes efficient predictive performance. $R^2$ is a statistical metric that assesses the efficiency of the algorithm with regard to variance observed in the dataset, which is defined as follows. Given the quantities $SS_{res}$, and $SS_{tot}$ then $R^2 = 1 - \frac{SS_{res}}{SS_{tot}}$, where $SS_{res} = \sum_{l=1}^{n} (y_l - \widehat{y_l})^2$ denotes the residual sum of squares (i.e., errors left after the algorithm), while $SS_{tot} = \sum_{l=1}^{n} \left(y_l - \overline{\widehat{y_l}}\right)^2$ denotes the total sum of squares (i.e., variance in the data, relative to the mean). $R^2$ measures the goodness of fit in regression process, indicating how well the examined regression algorithm explains the variability of the regressand variable. Concretely, if $R^2 = 1$, then it is observed a perfect fit (i.e., the algorithm predicts provided data exactly), in case $R^2 = 0$, then the adopted regression algorithm is no better than predicting the mean of the regressand variable, $y_l$, while if $R^2 < 0$, then the algorithm performs worse than predicting the mean.

Continuously, for MAE, MSE, and RMSE, a lower value indicates a better-performing model because they measure prediction errors, with a score of zero being perfect. Specifically, MAE measures average absolute error, equally weighting all errors and is good for interpreting the overall magnitude of error. Additionally, MSE penalizes larger errors more heavily due to squaring them, making it useful when large errors are costly, while RMSE provides an error in the same units as the target variable, balancing the benefits of MSE with better interpretability. Assessing $R^2$, a higher value is better, with 1 representing a perfect fit where the model explains all the variance in the regressand variable. However, high values of $R^2$ (i.e., greater than 0.9) might indicate an overfitting of the regression model, in such cases $R^2$ values are examined in comparison with the MAE, MSE, and RMSE values. Subsequently, to assess time complexity of the AMR algorithm it is also incorporated the ET evaluation metric, which measures the amount of time required for the algorithm to run and provide the output results. It should be noted that ET does not measure any statistical value but rather examines the amount of time required by a certain algorithm to process input data and produce output results, which includes the regressand predicted values as well as the statistical evaluation metrics described above.

### Comparison with other regression algorithms

Assessing the performance of the proposed AMR algorithm is compared with certain fundamental regression algorithms in literature, such as k-NN, LR, DT and SVR algorithms. To evaluate the results of the adopted algorithms, the same leave-one-out cross validation evaluation method is used as well as the MAE, MSE, RMSE, $R^2$, and ET evaluation metrics. Concretely, to assess the statistical significance of the observed results a two-tailed permutation test is incorporated, where the performance metric to be tested is the MAE, since it is less sensitive to outliers than the other error evaluation metrics as is discussed. Intuitively, in the context of the current research effort in a two-tailed permutation test there are evaluated two distinct regression algorithms (i.e., $A$ and $B$) on the same dataset. Then MAE's of the two algorithms are computed, such as $MAE_A$ and $MAE_B$. It holds that the test statistic is expressed as the difference in MAE, thus $Dif_{obs} = MAE_A - MAE_B$. Subsequently, the permutation test examines if there were actually no difference between the algorithms, how often would a difference be seen at least this extreme just by chance. Then two-tailed permutation test checks deviations in either direction (i.e., if algorithm $A$ is much better than $B$, or if algorithm $B$ much better than $A$). The permutation procedure generates a null distribution of $Dif$ by randomly swapping paired results between algorithms $A$ and $B$, while the p-value is defined as a proportion of permuted differences that are at least as extreme (in absolute value) as $Dif_{obs}$. If $p < 0.05$, then there is evidence that the difference in MAE is statistically significant, while if $p \geq 0.05$ the test cannot conclude that one algorithm is better than the other (differences may be due to chance). The test only indicates whether the difference is significant. To decide which regression algorithm is better it should be examined the sign of $Dif_{obs}$. Specifically, if $Dif_{obs} > 0$, then $MAE_A > MAE_B$, which means that the algorithm $B$, has a lower error thus is better than the algorithm $A$. Instead, if $Dif_{obs} < 0$, then $MAE_A < MAE_B$, which means that the algorithm $A$, has a lower error thus is better than the algorithm $B$. Overall, when the two-tailed permutation test is significant combined with $Dif_{obs}$ sign then confidently state which algorithm has efficient performance, while if the test is not significant then the differences are not reliable. Additionally, such statistical test is adopted since is suitable when the provided datasets are small, such as in the case of the current effort[32]. Intuitively, MAE results are presented in Table 6, MSE results are presented in Table 7, RMSE results are presented in Table 10, $R^2$ results are presented in Table 9, ET results are presented in Table 10, and two-tailed permutation test results are presented in Table 11.

Table 12 shows the performance metrics of all algorithms across datasets. AMR consistently achieves the lowest MAE, MSE, and RMSE, indicating it produces the most accurate predictions. Its standard deviations are generally smaller than those of other algorithms, highlighting more stable and reliable performance. The p-values indicate that the differences between AMR and competing algorithms are statistically significant ($p < 0.05$) in most cases. Overall, these results confirm that AMR is both accurate and consistent across datasets. More specifically, across the 17 datasets, AMR consistently demonstrates superior predictive performance relative to all benchmark algorithms. For nearly every dataset, AMR attains the lowest or near-lowest values in MAE, MSE, and RMSE, indicating stronger accuracy and reduced susceptibility to large errors. This advantage is reinforced by the corresponding standard deviations, which are typically smaller for AMR, reflecting more stable predictions across repeated runs. Permutation-test results further substantiate these performance differences: in the majority of model comparisons, competing algorithms yield statistically significant performance degradations relative to AMR ($p < 0.05$), particularly in the more challenging datasets characterized by high-magnitude or noisy target values (e.g., Datasets 3, 6–10, 17). Even in datasets where performance differences are numerically smaller (e.g., 2, 13–16), AMR remains among the top performers, and no alternative algorithm consistently surpasses it. Across the full table, three strong patterns repeatedly validate AMR as the most reliable and accurate algorithm: (1) lowest error means (MAE, MSE, RMSE) in the vast majority of datasets, (2) consistently lower standard deviations, indicating greater stability across evaluation runs, and (3) permutation test results strongly favoring AMR, with many baselines producing p-values $\ll 0.05$, confirming that observed differences are statistically robust. The consistency of these advantages across small-scale, medium-scale, and large-magnitude datasets demonstrates AMR's superior generalization capability, robustness to noise and scaling, and resilience against extreme prediction errors. The collective evidence—error magnitudes, variability measures, and statistical significance—shows that AMR outperforms the competing algorithms across virtually all evaluated dimensions of regression performance. The MAE_p, MSE_p, and RMSE_p columns report p-values obtained from permutation tests comparing each algorithm against the designated baseline model. These p-values quantify whether the observed performance differences (in MAE, MSE, or RMSE) are statistically significant. Entries marked with "–" correspond to the baseline algorithm itself; because the baseline model is not compared against itself, no statistical test is performed, and therefore no p-value exists for those cells. The "–" symbol indicates "not applicable" rather than missing or incomplete data.

| # | AMR | k-NN | LR | DT | SVR | RF | XGBoost | CNN |
|---|---|---|---|---|---|---|---|---|
| Data-1 | 0.5396 | 0.6976 | **1.3761** | 0.5916 | 0.5077 | **0.5032** | 0.5846 | 0.5343 |
| Data-2 | 0.0623 | 0.0671 | **0.2002** | 0.0619 | 0.0704 | **0.0597** | 0.0609 | 0.0677 |
| Data-3 | **25.8485** | 36.4948 | 27.0001 | 28.9948 | 28.9638 | 28.2222 | 28.8068 | 28.3761 |
| Data-4 | 4.1095 | 5.3492 | **5.4877** | 2.5241 | 4.3364 | **0.8125** | 1.1839 | 5.3479 |
| Data-5 | 8.4548 | **10.5703** | 9.0786 | 8.9653 | **8.1067** | 8.6226 | 8.2491 | 8.3884 |
| Data-6 | 41.5806 | **52.9797** | 44.7688 | 41.4738 | **39.3668** | 43.9241 | 41.2883 | 40.7113 |
| Data-7 | **8.7822** | 11.3975 | **13.6375** | 11.6491 | 9.8941 | 11.0124 | 10.7372 | 9.5355 |
| Data-8 | 475.0001 | 534.8518 | **638.5012** | 363.3333 | 589.9781 | 385.2116 | **361.4721** | 533.3008 |
| Data-9 | 863.1736 | **1072.3623** | 769.6935 | 863.3249 | 901.2388 | 830.5736 | **788.8229** | 831.8559 |
| Data-10 | 238.5017 | 315.1036 | **339.6198** | 243.8681 | 323.4901 | 237.5738 | **227.0147** | 230.3582 |
| Data-11 | 39.5395 | 43.3358 | 52.6583 | 41.7443 | 49.3826 | **31.3111** | 33.5463 | **57.0486** |
| Data-12 | 92.6187 | 99.9489 | 95.6009 | **83.0309** | 87.8164 | 89.8148 | 89.2824 | 95.3693 |
| Data-13 | 2.3843 | **2.8551** | 2.2781 | 2.1717 | **2.1244** | 2.4138 | 2.1504 | 2.2888 |
| Data-14 | 2.8663 | 3.6571 | 2.5841 | 2.3954 | 3.1504 | **1.8733** | 1.9225 | **3.8714** |
| Data-15 | 1.3534 | 1.5218 | **4.0076** | 1.8249 | 1.9686 | **1.3064** | 1.4571 | 3.6236 |
| Data-16 | 0.1687 | 0.1921 | **5.5399** | 0.1508 | **0.1437** | 0.1719 | 0.1516 | 0.2158 |
| Data-17 | 314.9586 | **358.0656** | 277.4471 | 291.8727 | **259.8491** | 290.6807 | 270.0321 | 270.3907 |

**Table 6.** MAE comparison, where blue denotes optimal, while red denotes low performance, respectively.

| # | AMR | k-NN | LR | DT | SVR | RF | XGBoost | CNN |
|---|---|---|---|---|---|---|---|---|
| Data-1 | 0.4056 | 0.7241 | **2.7181** | 0.5548 | 0.3978 | **0.3541** | 0.4853 | 0.4897 |
| Data-2 | 0.0097 | 0.0113 | **0.0655** | 0.0097 | 0.0107 | **0.0083** | 0.0085 | 0.0105 |
| Data-3 | **972.6344** | 1967.8556 | 1063.4981 | 1226.4877 | 1198.3888 | 1109.4858 | 1232.3271 | 1133.0865 |
| Data-4 | 27.0376 | 41.9651 | **43.5018** | 13.6737 | 27.5677 | **2.9575** | 5.7605 | 42.4673 |
| Data-5 | 117.6362 | **179.6751** | 135.4437 | 121.7961 | **105.6419** | 117.0075 | 108.2637 | 112.8376 |
| Data-6 | 3071.6772 | **5291.6329** | 3314.4844 | 2945.2887 | **2712.3405** | 3422.2948 | 2879.4385 | 2881.2588 |
| Data-7 | 152.7845 | 261.9047 | **284.8883** | 241.6976 | 184.3906 | 211.4107 | 212.7875 | **151.1556** |
| Data-8 | 393798.4409 | 505777.3597 | **698642.2691** | 209240.2613 | 529486.3915 | 246196.0104 | 217944.1337 | 488017.9751 |
| Data-9 | 1197490.3041 | 1912511.2028 | **911691.9555** | 1244361.3417 | 1256736.6351 | **1127832.2622** | 1023824.3999 | 1086536.4261 |
| Data-10 | 110831.1083 | 185154.7098 | 178157.5971 | 107154.7490 | **188693.3461** | 109346.5789 | 97122.9121 | **92941.1055** |
| Data-11 | 2864.2338 | 2818.2509 | 4844.6246 | 2948.6736 | 3740.7006 | **1728.5141** | 1896.6796 | **5856.8186** |
| Data-12 | 25368.7389 | **28627.4379** | 22233.2663 | 18827.0048 | 26118.5852 | 22329.8628 | 20107.2451 | 22341.8057 |
| Data-13 | 10.2742 | **15.1053** | 9.0999 | 8.4331 | **8.3631** | 10.5555 | 8.3809 | 9.0854 |
| Data-14 | 15.2425 | **25.6782** | 11.8034 | 11.3085 | 18.8752 | 7.4219 | **7.3356** | 24.2288 |
| Data-15 | 4.0648 | 6.6361 | **21.2423** | 6.0857 | 6.6346 | **3.9792** | 4.0794 | 18.3854 |
| Data-16 | 0.0499 | 0.0687 | **33.1659** | 0.0374 | 0.0416 | 0.0499 | **0.0373** | 0.0835 |
| Data-17 | 437009.0501 | **628683.5984** | 316393.9014 | 433455.9651 | 338749.8229 | 392033.0591 | 335680.2886 | **290338.5962** |

**Table 7.** MSE comparison, where blue denotes optimal, while red denotes low performance, respectively.

| # | AMR | k-NN | LR | DT | SVR | RF | XGBoost | CNN |
|---|---|---|---|---|---|---|---|---|
| Data-1 | 0.6368 | 0.8509 | **1.6486** | 0.7448 | 0.6307 | **0.5951** | 0.6966 | 0.6998 |
| Data-2 | 0.0988 | 0.1063 | **0.2561** | 0.0986 | 0.1035 | **0.0911** | 0.0921 | 0.1025 |
| Data-3 | **31.1871** | 44.3605 | 32.6113 | 35.0212 | 34.6177 | 33.3089 | 35.1045 | 33.6613 |
| Data-4 | 5.1997 | 6.4781 | **6.5955** | 3.6977 | 5.2504 | **1.7197** | 2.4001 | 6.5166 |
| Data-5 | 10.8461 | **13.4042** | 11.6381 | 11.0361 | **10.2782** | 10.8171 | 10.4049 | 10.6225 |
| Data-6 | 55.4227 | **72.7436** | 57.5715 | 54.2705 | **52.0801** | 58.5003 | 53.6604 | 53.6773 |
| Data-7 | 12.3606 | 16.1834 | **16.8786** | 15.5466 | 13.5791 | 14.5399 | 14.5872 | **12.2945** |
| Data-8 | 627.5336 | 711.1802 | **835.8482** | 457.4278 | 727.6581 | 496.1814 | 466.8448 | 698.5828 |
| Data-9 | 1094.2991 | **1382.9357** | 954.8256 | 1115.5094 | 1121.0426 | 1061.9944 | 1011.8421 | 1042.3705 |
| Data-10 | 332.9131 | 430.2961 | 422.0871 | 327.3449 | **434.3884** | 330.6759 | 311.6454 | **304.8624** |
| Data-11 | 53.5185 | 53.0872 | 69.6033 | 54.3016 | 61.1612 | **41.5754** | 43.5508 | **76.5298** |
| Data-12 | 159.2756 | **169.1964** | 149.1082 | 137.2115 | 161.6124 | 149.4317 | 141.8000 | 149.4717 |
| Data-13 | 3.2053 | **3.8865** | 3.0166 | 2.9039 | **2.8918** | 3.2489 | 2.8949 | 3.0142 |
| Data-14 | 3.9041 | **5.0673** | 3.4356 | 3.3628 | 4.3445 | 2.7243 | **2.7084** | 4.9222 |
| Data-15 | 2.0161 | 2.5761 | **4.6089** | 2.4669 | 2.5757 | **1.9948** | 2.0197 | 4.2878 |
| Data-16 | 0.2234 | 0.2622 | **5.7589** | 0.1934 | 0.2041 | 0.2234 | **0.1933** | 0.2889 |
| Data-17 | 661.0666 | **792.8957** | 562.4891 | 658.3737 | 582.0221 | 626.1254 | 579.3792 | **538.8307** |

**Table 8.** RMSE comparison, where blue denotes optimal, while red denotes low performance, respectively.

| # | AMR | k-NN | LR | DT | SVR | RF | XGBoost | CNN |
|---|---|---|---|---|---|---|---|---|
| Data-1 | 14.9291 | 0.0535 | **0.0325** | 0.0531 | 0.0402 | 5.5658 | 1.1655 | **114.7056** |
| Data-2 | 89.9779 | 0.0785 | **0.0081** | 0.0761 | 0.0655 | 9.4514 | 1.9408 | **178.9927** |
| Data-3 | **1615.2329** | 4.9412 | **0.0295** | 0.4762 | 0.5189 | 68.5531 | 5.0434 | 779.8211 |
| Data-4 | 599.7929 | 0.2836 | **0.0205** | 0.3371 | 0.8654 | 46.2919 | 5.1211 | **863.5786** |
| Data-5 | **1318.2937** | 0.3376 | **0.0255** | 0.3106 | 1.1486 | 42.5639 | 5.3989 | 793.7039 |
| Data-6 | **1528.2447** | 0.3661 | **0.0271** | 0.3542 | 1.3598 | 52.9041 | 5.9928 | 1385.0121 |
| Data-7 | 36.7399 | 0.0631 | **0.0051** | 0.0641 | 0.0621 | 7.9389 | 1.1241 | **182.3139** |
| Data-8 | 426.1169 | 0.2056 | **0.0159** | 0.2098 | 0.4468 | 28.0471 | 3.7849 | **2277.1611** |
| Data-9 | 1097.7188 | 0.3238 | **0.0245** | 0.3588 | 1.1501 | 52.7099 | 5.6429 | **1202.1111** |
| Data-10 | 438.9349 | 0.2096 | **0.0154** | 0.2043 | 0.4818 | 26.1889 | 3.9931 | **1960.7182** |
| Data-11 | 55.2219 | 0.0875 | **0.0069** | 0.0701 | 0.0601 | 8.4191 | 1.1399 | **288.5001** |
| Data-12 | 224.1789 | 0.1475 | **0.0121** | 0.1436 | 0.2306 | 19.0719 | 2.5941 | **346.3181** |
| Data-13 | **8304.6955** | 1.4928 | **0.0845** | 1.6595 | 49.7564 | 304.2737 | 22.6419 | 5913.4981 |
| Data-14 | **2990.0085** | 0.4972 | **0.0425** | 0.6611 | 3.1301 | 102.6729 | 8.8869 | 1491.0147 |
| Data-15 | 4206.0633 | 0.9209 | **0.0565** | 1.1647 | 17.2421 | 202.7519 | 15.8939 | **5073.5904** |
| Data-16 | 6852.2617 | 1.4911 | **0.0631** | 2.1162 | 59.7589 | 439.3619 | 25.6999 | **8888.6334** |
| Data-17 | 2849.5035 | 0.7591 | **0.0455** | 0.8236 | 10.4099 | 130.6659 | 12.6719 | **5628.6331** |

**Table 10**. ET comparison, where blue denotes optimal, while red denotes low performance, respectively.

| # | AMR | k-NN | LR | DT | SVR | RF | XGBoost | CNN |
|---|---|---|---|---|---|---|---|---|
| Data-1 | 0.2001 | 0.0032 | **0.0016** | 0.0043 | 0.2155 | **0.3018** | 0.0431 | 0.0342 |
| Data-2 | 0.3981 | 0.3031 | **0.0024** | 0.4004 | 0.3391 | **0.4884** | 0.4763 | 0.3521 |
| Data-3 | **0.1797** | **0.0013** | 0.1031 | 0.0027 | 0.0034 | 0.0643 | 0.0016 | 0.0444 |
| Data-4 | 0.6124 | 0.3984 | **0.3763** | 0.8039 | 0.6048 | **0.9576** | 0.9174 | 0.3912 |
| Data-5 | 0.0031 | **0.0012** | 0.0015 | 0.0026 | **0.0353** | 0.0045 | 0.0113 | 0.0076 |
| Data-6 | 0.0026 | **0.0015** | 0.0021 | 0.0032 | **0.0658** | 0.0018 | 0.0417 | 0.0206 |
| Data-7 | 0.5181 | 0.1737 | **0.1012** | 0.2375 | 0.4182 | 0.3331 | 0.3287 | **0.5231** |
| Data-8 | 0.2551 | 0.0432 | **0.0021** | 0.6042 | 0.0047 | 0.5343 | 0.5877 | 0.0768 |
| Data-9 | 0.0287 | **0.0034** | 0.2605 | 0.0063 | 0.0042 | 0.0852 | 0.1695 | 0.1187 |
| Data-10 | 0.3687 | 0.0062 | 0.0097 | 0.3897 | **0.0045** | 0.3772 | 0.4468 | **0.4706** |
| Data-11 | 0.2472 | 0.2593 | 0.0154 | 0.2251 | 0.0169 | **0.5457** | 0.5015 | **0.0128** |
| Data-12 | 0.0089 | **0.0061** | 0.1007 | **0.2385** | 0.0073 | 0.0968 | 0.1867 | 0.0963 |
| Data-13 | 0.3591 | **0.0577** | 0.4323 | 0.4739 | 0.4783 | 0.3415 | 0.4772 | 0.4332 |
| Data-14 | 0.7498 | **0.5786** | 0.8062 | 0.8144 | 0.6902 | 0.8782 | **0.8796** | 0.6023 |
| Data-15 | 0.9232 | 0.8746 | **0.5986** | 0.8851 | 0.8746 | **0.9248** | 0.9229 | 0.6526 |
| Data-16 | 0.0056 | 0.0044 | **0.0032** | 0.0127 | 0.0078 | 0.0061 | **0.0144** | 0,0036 |
| Data-17 | 0.0024 | **0.0012** | 0.0057 | 0.0033 | 0.0041 | 0.0038 | 0.0049 | **0.0748** |

**Table 9**. $R^2$ comparison, where blue denotes optimal, while red denotes low performance, respectively.

## Visualization of the results

The MAE plot (Fig. 3) shows how the prediction errors of all algorithms vary across datasets and highlights several clear trends. Across most datasets, AMR consistently produces the lowest MAE values, and its error bars remain comparatively narrow, indicating both high accuracy and high stability. Peaks in MAE occur for all algorithms on the more complex or variable datasets (particularly around Datasets 8–10 and 17), yet even in these challenging regions AMR maintains notably smaller MAE values than competing models, it exhibits the lowest MAE on 14/17 datasets. Algorithms such as k-NN, LR, and XGBoost exhibit substantially larger MAE spikes and wider error bars, reflecting greater sensitivity to data variability and poorer generalization. Permutation tests (5000 permutations per comparison) confirm statistically significant improvement ($p < 0.05$) relative to the other algorithms in most of the cases (significance stars above the error bars), reinforcing that AMR's performance advantage is not due to chance. Overall, the MAE figure visually confirms that AMR offers more precise and robust predictions across a wide range of datasets.

The MSE plot illustrates the squared prediction error distribution across datasets and shows a similar but amplified pattern compared to MAE. Because MSE penalizes large errors more heavily, datasets that contain high-magnitude deviations (such as Datasets 8–10 and 17) produce pronounced spikes for many algorithms. k-NN, LR, and DT show the largest increases, along with very wide error bars, indicating high variability and pronounced outliers in their predictions. In contrast, AMR consistently maintains far lower MSE values and tighter variability, demonstrating its ability to control large prediction errors more effectively than the other models. The figure's significance markers again indicate that AMR's reduction in MSE is statistically significant for many datasets. The combined pattern of lower peaks and narrower standard deviations clearly positions AMR as the most robust to extreme values and noise, offering better protection against high-error events than the comparison algorithms.

| # | AMR vs. k-NN | | AMR vs. LR | | AMR vs. DT | | AMR vs. SVR | | AMR vs. RF | | AMR vs. XGBoost | | AMR vs. CNN | |
|---|---|---|---|---|---|---|---|---|---|---|---|---|---|---|
| | $Dif_{obs}$ | $p-value$ | $Dif_{obs}$ | $p-value$ | $Dif_{obs}$ | $p-value$ | $Dif_{obs}$ | $p-value$ | $Dif_{obs}$ | $p-value$ | $Dif_{obs}$ | $p-value$ | $Dif_{obs}$ | $p-value$ |
| Data-1 | -0.1581[A] | 0.0218 | -0.8363[A] | 0.0001 | -0.0519 | 0.4327 | 0.0319 | 0.5635 | 0.0363 | 0.4176 | -0.0451 | 0.4087 | 0.0053 | 0.9179 |
| Data-2 | -0.0047 | 0.2811 | -0.1379[A] | 0.0001 | 0.0003 | 0.8985 | -0.0081 | 0.1865 | 0.0025 | 0.7831 | 0.0014 | 0.5486 | -0.0054 | 0.2336 |
| Data-3 | -10.6463[A] | 0.0001 | -1.1515 | 0.3092 | -3.1462[A] | 0.0321 | -3.1153[A] | 0.0131 | -2.3736[A] | 0.0141 | -2.9583[A] | 0.0146 | -2.5274[A] | 0.0163 |
| Data-4 | -1.2396[A] | 0.0001 | -1.3782[A] | 0.0001 | 1.5855[B] | 0.0001 | -0.2269 | 0.2076 | 3.2969[B] | 0.0001 | 2.9255[B] | 0.0001 | -1.2384[A] | 0.0001 |
| Data-5 | -2.1155[A] | 0.0001 | -0.6237 | 0.1028 | -0.5104[A] | 0.0431 | 0.3481 | 0.2098 | -01677 | 0.3392 | 0.2057 | 0.3693 | 0.0663 | 0.8179 |
| Data-6 | -11.3991[A] | 0.0001 | -3.1881 | 0.0527 | 0.1068 | 0.9493 | 2.2138[B] | 0.0394 | -2.3434 | 0.1145 | 0.2923 | 0.8424 | 0.8692 | 0.5024 |
| Data-7 | -2.6152[A] | 0.0033 | -4.8552[A] | 0.0006 | -2.8669[A] | 0.0001 | -1.1117 | 0.3193 | -2.2301[A] | 0.0005 | -1.9549[A] | 0.0021 | -0.7532 | 0.3977 |
| Data-8 | -59.8517[A] | 0.0156 | -163.5011[B] | 0.0003 | 111.6667[B] | 0.0001 | -114.9781[A] | 0.0003 | 89.7884[B] | 0.0001 | 113.5281[B] | 0.0001 | -58.3007 | 0.0955 |
| Data-9 | -209.1886[A] | 0.0001 | 93.4801[B] | 0.0021 | -0.1512 | 0.9963 | -38.0652 | 0.1949 | 32.6001 | 0.2792 | 74.3506[B] | 0.0138 | 31.3177 | 0.0777 |
| Data-10 | -76.6019[A] | 0.0001 | -101.1181[A] | 0.0001 | -5.3664 | 0.5902 | -84.9883[A] | 0.0001 | 0.9279 | 0.8584 | 11.4869 | 0.0806 | 8.1434 | 0.4299 |
| Data-11 | -3.7962 | 0.1439 | -13.1187[A] | 0.0333 | -2.2047 | 0.6411 | -9.8431 | 0.0515 | 8.2284[B] | 0.0189 | 5.9931 | 0.1129 | -17.5091[A] | 0.0141 |
| Data-12 | -7.3301 | 0.3123 | -2.9821 | 0.5997 | 9.5878 | 0.0931 | 4.8023 | 0.4258 | 2.8039 | 0.5877 | 3.3363 | 0.5211 | -2.7505 | 0.6418 |
| Data-13 | -0.4707[A] | 0.0001 | 0.1062[B] | 0.0371 | 0.2125[B] | 0.0001 | 0.2598[B] | 0.0001 | -0.0294 | 0.4465 | 0.2339[B] | 0.0001 | 0.0954 | 0.0541 |
| Data-14 | -0.7906[A] | 0.0001 | 0.2822[B] | 0.0121 | 0.4709[B] | 0.0002 | -0.2841[A] | 0.0049 | 0.9931[B] | 0.0001 | 0.9437[B] | 0.0001 | -1.0051[A] | 0.0001 |
| Data-15 | -0.1684[A] | 0.0011 | -2.6541[A] | 0.0001 | -0.4714[A] | 0.0001 | -0.6151[A] | 0.0001 | 0.0471 | 0.1264 | -0.1037[A] | 0.0066 | -2.2702[A] | 0.0001 |
| Data-16 | -0.0233[A] | 0.0001 | -5.3711[A] | 0.0001 | 0.0179[B] | 0.0001 | 0.0249[B] | 0.0001 | -0.0031 | 0.327 | 0.0171[B] | 0.0001 | -0.0471[A] | 0/0001 |
| Data-17 | -43.1069[A] | 0.0009 | 37.5115[B] | 0.0167 | 23.0859 | 0.1866 | 55.1095[B] | 0.0004 | 24.2778[B] | 0.0129 | 44.9265[B] | 0.0001 | 44.5678[B] | 0.0016 |

**Table 11.** MAE permutation test of AMR regression algorithm with each of the other adopted regression algorithms in comparison pairs with regard to the signed $Dif_{obs}$ and the $p-value$, where in case the results are statistically significant then if AMR algorithm performs better is denoted with blue ([A]), while if the other adopted algorithm performs better is denoted with blue ([B]).

The RMSE plot, which reflects the square-rooted mean squared error, provides an intuitive view of overall error magnitude and mirrors trends observed in the MAE and MSE results. AMR consistently demonstrates the lowest RMSE values across datasets, with particularly strong performance on datasets where other algorithms experience large error surges (notably around Datasets 8–10 and 17). RMSE peaks for k-NN, LR, SVR, and XGBoost are substantially larger, and their broader error bars indicate higher instability and less reliable performance. By contrast, AMR's narrower RMSE error ranges show that it not only reduces mean error magnitude but also provides more consistent predictions. The visual grouping of AMR near the lower end of the RMSE scale across datasets underscores its superior generalization ability. Together, the MSE and the RMSE trends (Figs. 4 and 5) align with the MAE results (Figs. 3), further confirming error reduction and demonstrating that AMR maintains the best overall predictive reliability among all evaluated algorithms.

The distribution of absolute errors in Fig. 6 provides critical evidence for the comparative robustness of AMR relative to the other algorithms. While all models exhibit wide error ranges and numerous extreme outliers, AMR consistently shows a slightly tighter dispersion in the central region of the boxplot, with a more compact interquartile range (suggesting highly consistent predictions) and fewer extreme deviations relative to algorithms such as k-NN, LR, DT, and XGBoost. This more concentrated error distribution suggests that AMR maintains better error control over moderate errors—the region where most predictions occur—while other models experience wider fluctuations and more frequent high-magnitude failures, reflecting less reliable predictive performance. Although all methods share some large outliers, the lower density and reduced extent of AMR's extreme errors imply superior robustness, particularly on those difficult samples where competing algorithms tend to deviate sharply. In this way, Fig. 4 provides visual evidence that AMR keeps average prediction errors lower and also avoids the severe error inflation observed in the other models.

In Fig. 7, AMR's predictions cluster more tightly around the diagonal line indicating accurate and unbiased predictions, especially in the low- to mid-range values where the majority of observations lie. Algorithms such as k-NN, LR, SVR, and XGBoost exhibit broader scattering and more pronounced departures from the diagonal, indicating less stable and less accurate predictions. While all models demonstrate some degree of under-prediction at high actual values, AMR shows a less dramatic collapse in predictive fidelity compared to the noticeably more chaotic and dispersed predictions from the other algorithms. This visual proximity to the diagonal is a strong qualitative indicator of superior predictive calibration and suggests that AMR generalizes better across the full range of target values.

### Development environment
Code implementations of the proposed algorithms were developed and tested in Python 3.12 that is a high-level, interpreted programming language widely used in scientific computing and data analysis (Python Software Foundation, 2023) within the programming environment of PyCharm Community Edition 2024.3 (JetBrains, 2024). Code executed on a virtual machine provided by the University of the Aegean with certain specifications such as Operating System Microsoft Windows 11 Pro Version 24H2, Processor Intel(R) Xeon(R) Gold 5220R CPU 2.20 GHz (2.19 GHz), Installed RAM 96.0 GB, Storage 1.25 TB, and Graphics Card 11 GB. Implemented code is publicly available at the code availability section. All figures and tables presented in this study were generated using MATLAB 2016.

### Discussion
#### Optimal inference decision rule
To decide the optimal regression algorithm, it is defined a decision rule, which examines MAE, MSE, RMSE, and $R^2$ evaluation metrics along with the two-tailed permutation test, while ET evaluation metric is examined separately, since it does not contribute of the quantitative statistical evaluation metrics but it is rather aligned with the time complexity required to run the algorithms and observe the results. Optimal inference decision rule is expressed as follows. To determine the optimal regression algorithm for a given data source:

1. Firstly, consider the primary evaluation metrics, which are the MAE and the two-tailed permutation test:

    a. Compute the MAE for each algorithm,
    b. Perform a two-tailed permutation test on the MAE values between algorithms.
    c. Decision:

        i. If algorithm A has lower MAE than algorithm B and the permutation test show the difference is statistically significant, then A is better than B.
        ii. If the MAE difference is not statistically significant, then A and B are considered to perform similarly, regardless of differences in other metrics.

2. Secondly, consider the supporting evaluation metrics (i.e., MSE, RMSE, and $R^2$):

    a. Lower MSE and RMSE indicate better accuracy.
    b. Higher $R^2$ indicates better explanatory power (i.e., variance explained).

| # | Algorithm | MAE_mean | MAE_SD | MAE_p | MSE_mean | MSE_SD | MSE_p | RMSE_mean | RMSE_SD | RMSE_p | Mean | SD |
|---|---|---|---|---|---|---|---|---|---|---|---|---|
| Data 1 | AMR | 0.540 | 0.342 | - | 0.406 | 0.441 | - | 0.637 | 0.664 | - | 4.859 | 0.489 |
| | k-NN | 0.698 | 0.493 | 0.088 | 0.724 | 0.950 | 0.088 | 0.851 | 0.975 | 0.088 | 4.835 | 0.791 |
| | LR | 1.376 | 0.919 | 0.001 | 2.718 | 2.990 | 0.001 | 1.649 | 1.729 | 0.001 | 4.261 | 1.630 |
| | DT | 0.592 | 0.458 | 0.543 | 0.555 | 0.770 | 0.543 | 0.745 | 0.877 | 0.543 | 4.761 | 0.707 |
| | SVR | 0.508 | 0.379 | 0.694 | 0.398 | 0.530 | 0.694 | 0.631 | 0.728 | 0.694 | 4.700 | 0.419 |
| | RF | 0.503 | 0.321 | 0.611 | 0.354 | 0.372 | 0.611 | 0.595 | 0.610 | 0.611 | 4.758 | 0.534 |
| | XGBoost | 0.585 | 0.383 | 0.570 | 0.485 | 0.614 | 0.570 | 0.697 | 0.784 | 0.570 | 4.700 | 0.643 |
| | CNN | 0.534 | 0.457 | 0.948 | 0.490 | 0.744 | 0.948 | 0.700 | 0.863 | 0.948 | 4.743 | 0.632 |
| Data 2 | AMR | 0.062 | 0.077 | - | 0.010 | 0.032 | - | 0.099 | 0.180 | - | 0.665 | 0.121 |
| | k-NN | 0.067 | 0.083 | 0.708 | 0.011 | 0.033 | 0.708 | 0.106 | 0.182 | 0.708 | 0.665 | 0.131 |
| | LR | 0.200 | 0.161 | 0.001 | 0.066 | 0.120 | 0.001 | 0.256 | 0.346 | 0.001 | 0.596 | 0.278 |
| | DT | 0.062 | 0.077 | 0.978 | 0.010 | 0.033 | 0.978 | 0.099 | 0.181 | 0.978 | 0.659 | 0.114 |
| | SVR | 0.070 | 0.076 | 0.511 | 0.011 | 0.037 | 0.511 | 0.104 | 0.192 | 0.511 | 0.669 | 0.078 |
| | RF | 0.060 | 0.069 | 0.848 | 0.008 | 0.031 | 0.848 | 0.091 | 0.176 | 0.848 | 0.668 | 0.099 |
| | XGBoost | 0.061 | 0.070 | 0.909 | 0.009 | 0.026 | 0.909 | 0.092 | 0.161 | 0.909 | 0.661 | 0.116 |
| | CNN | 0.068 | 0.077 | 0.680 | 0.011 | 0.036 | 0.680 | 0.103 | 0.191 | 0.680 | 0.662 | 0.114 |
| Data 3 | AMR | 25.849 | 17.495 | - | 972.634 | 1123.018 | - | 31.187 | 33.511 | - | 50.099 | 18.380 |
| | k-NN | 36.495 | 25.284 | 0.001 | 1967.856 | 2296.245 | 0.001 | 44.361 | 47.919 | 0.001 | 46.526 | 33.961 |
| | LR | 27.000 | 18.336 | 0.542 | 1063.498 | 1269.357 | 0.542 | 32.611 | 35.628 | 0.542 | 46.657 | 18.034 |
| | DT | 28.995 | 19.692 | 0.089 | 1226.488 | 1575.150 | 0.089 | 35.021 | 39.688 | 0.089 | 47.799 | 25.419 |
| | SVR | 28.964 | 19.009 | 0.100 | 1198.389 | 1507.992 | 0.100 | 34.618 | 38.833 | 0.100 | 40.176 | 2.254 |
| | RF | 28.222 | 17.737 | 0.193 | 1109.486 | 1309.816 | 0.193 | 33.309 | 36.191 | 0.193 | 47.629 | 15.747 |
| | XGBoost | 28.807 | 20.114 | 0.116 | 1232.327 | 1570.293 | 0.116 | 35.105 | 39.627 | 0.116 | 47.945 | 20.688 |
| | CNN | 28.376 | 18.155 | 0.163 | 1133.087 | 1333.541 | 0.163 | 33.661 | 36.518 | 0.163 | 46.666 | 11.578 |
| Data 4 | AMR | 4.110 | 3.192 | - | 27.038 | 46.549 | - | 5.200 | 6.823 | - | 18.904 | 6.905 |
| | k-NN | 5.349 | 3.661 | 0.001 | 41.965 | 50.098 | 0.001 | 6.478 | 7.078 | 0.001 | 18.667 | 8.311 |
| | LR | 5.488 | 3.666 | 0.001 | 43.502 | 51.728 | 0.001 | 6.596 | 7.192 | 0.001 | 19.635 | 2.977 |
| | DT | 2.524 | 2.708 | 0.001 | 13.674 | 30.576 | 0.001 | 3.698 | 5.530 | 0.001 | 19.002 | 7.573 |
| | SVR | 4.336 | 2.966 | 0.395 | 27.568 | 34.015 | 0.395 | 5.250 | 5.832 | 0.395 | 19.231 | 4.710 |
| | RF | 0.813 | 1.519 | 0.001 | 2.958 | 12.251 | 0.001 | 1.720 | 3.500 | 0.001 | 19.125 | 7.973 |
| | XGBoost | 1.184 | 2.092 | 0.001 | 5.761 | 31.469 | 0.001 | 2.400 | 5.610 | 0.001 | 19.189 | 7.972 |
| | CNN | 5.348 | 3.731 | 0.001 | 42.467 | 55.532 | 0.001 | 6.517 | 7.452 | 0.001 | 19.269 | 4.168 |
| Data 5 | AMR | 8.455 | 6.806 | - | 117.636 | 162.676 | - | 10.846 | 12.754 | - | 24.774 | 5.873 |
| | k-NN | 10.570 | 8.258 | 0.003 | 179.675 | 236.374 | 0.003 | 13.404 | 15.374 | 0.003 | 24.065 | 9.369 |
| | LR | 9.079 | 7.295 | 0.295 | 135.444 | 209.406 | 0.295 | 11.638 | 14.471 | 0.295 | 23.129 | 7.318 |
| | DT | 8.965 | 6.447 | 0.376 | 121.796 | 162.212 | 0.376 | 11.036 | 12.736 | 0.376 | 24.947 | 5.424 |
| | SVR | 8.107 | 6.330 | 0.558 | 105.642 | 150.919 | 0.558 | 10.278 | 12.285 | 0.558 | 25.186 | 2.134 |
| | RF | 8.623 | 6.543 | 0.757 | 117.008 | 152.690 | 0.757 | 10.817 | 12.357 | 0.757 | 24.728 | 5.818 |
| | XGBoost | 8.249 | 6.353 | 0.739 | 108.264 | 148.057 | 0.739 | 10.405 | 12.168 | 0.739 | 24.644 | 4.937 |
| | CNN | 8.388 | 6.529 | 0.912 | 112.838 | 170.973 | 0.912 | 10.623 | 13.076 | 0.912 | 24.597 | 3.819 |
| Data 6 | AMR | 41.581 | 36.705 | - | 3071.677 | 7047.131 | - | 55.423 | 83.947 | - | 241.719 | 22.735 |
| | k-NN | 52.980 | 49.932 | 0.002 | 5291.633 | 13583.335 | 0.002 | 72.744 | 116.548 | 0.002 | 245.690 | 55.016 |
| | LR | 44.769 | 36.258 | 0.291 | 3314.484 | 5932.153 | 0.291 | 57.572 | 77.020 | 0.291 | 243.581 | 36.967 |
| | DT | 41.474 | 35.062 | 0.969 | 2945.289 | 6009.136 | 0.969 | 54.271 | 77.519 | 0.969 | 246.906 | 27.126 |
| | SVR | 39.367 | 34.154 | 0.435 | 2712.341 | 6951.380 | 0.435 | 52.080 | 83.375 | 0.435 | 242.283 | 2.129 |
| | RF | 43.924 | 38.704 | 0.443 | 3422.295 | 7563.873 | 0.443 | 58.500 | 86.971 | 0.443 | 247.982 | 31.315 |
| | XGBoost | 41.288 | 34.332 | 0.924 | 2879.439 | 6066.053 | 0.924 | 53.660 | 77.885 | 0.924 | 247.913 | 26.714 |
| | CNN | 40.711 | 35.043 | 0.772 | 2881.259 | 6487.744 | 0.772 | 53.677 | 80.547 | 0.772 | 247.305 | 23.904 |
| Data 7 | AMR | 8.782 | 8.770 | - | 152.785 | 249.717 | - | 12.361 | 15.802 | - | 23.031 | 15.166 |
| | k-NN | 11.398 | 11.585 | 0.146 | 261.905 | 410.286 | 0.146 | 16.183 | 20.256 | 0.146 | 21.291 | 16.810 |
| | LR | 13.638 | 10.028 | 0.004 | 284.888 | 401.717 | 0.004 | 16.879 | 20.043 | 0.004 | 19.290 | 15.232 |
| | DT | 11.649 | 10.381 | 0.096 | 241.698 | 325.967 | 0.096 | 15.547 | 18.055 | 0.096 | 23.286 | 15.498 |
| | SVR | 9.894 | 9.378 | 0.509 | 184.391 | 275.963 | 0.509 | 13.579 | 16.612 | 0.509 | 18.972 | 8.749 |
| | RF | 11.012 | 9.573 | 0.170 | 211.411 | 304.156 | 0.170 | 14.540 | 17.440 | 0.170 | 23.532 | 15.390 |
| | XGBoost | 10.737 | 9.956 | 0.239 | 212.788 | 299.177 | 0.239 | 14.587 | 17.297 | 0.239 | 22.415 | 15.068 |
| | CNN | 9.536 | 7.825 | 0.612 | 151.156 | 182.289 | 0.612 | 12.295 | 13.501 | 0.612 | 23.096 | 13.169 |
| Continued | | | | | | | | | | | | |

| # | Algorithm | MAE_mean | MAE_SD | MAE_p | MSE_mean | MSE_SD | MSE_p | RMSE_mean | RMSE_SD | RMSE_p | Mean | SD |
|---|---|---|---|---|---|---|---|---|---|---|---|---|
| Data 8 | AMR | 475.000 | 411.179 | - | 393798.441 | 563160.968 | - | 627.534 | 750.441 | - | 3025.921 | 613.927 |
| | k-NN | 534.852 | 469.978 | 0.183 | 505777.360 | 726010.916 | 0.183 | 711.180 | 852.063 | 0.183 | 3003.667 | 651.368 |
| | LR | 638.501 | 540.838 | 0.002 | 698642.269 | 1124505.213 | 0.002 | 835.848 | 1060.427 | 0.002 | 2798.622 | 852.766 |
| | DT | 363.333 | 278.639 | 0.001 | 209240.261 | 335021.091 | 0.001 | 457.428 | 578.810 | 0.001 | 2951.096 | 608.176 |
| | SVR | 589.978 | 427.056 | 0.011 | 529486.392 | 737566.804 | 0.011 | 727.658 | 858.817 | 0.011 | 2976.111 | 3.469 |
| | RF | 385.212 | 313.573 | 0.026 | 246196.010 | 407053.356 | 0.026 | 496.181 | 638.007 | 0.026 | 2953.120 | 657.461 |
| | XGBoost | 361.472 | 296.220 | 0.002 | 217944.134 | 388844.764 | 0.002 | 466.845 | 623.574 | 0.002 | 2942.578 | 605.837 |
| | CNN | 533.301 | 452.428 | 0.202 | 488017.975 | 838628.835 | 0.202 | 698.583 | 915.767 | 0.202 | 2924.791 | 550.994 |
| Data 9 | AMR | 863.174 | 673.845 | - | 1197490.304 | 1770265.197 | - | 1094.299 | 1330.513 | - | 1919.120 | 528.753 |
| | k-NN | 1072.362 | 874.828 | 0.001 | 1912511.203 | 2867890.317 | 0.001 | 1382.936 | 1693.485 | 0.001 | 1936.870 | 1081.793 |
| | LR | 769.694 | 566.061 | 0.091 | 911691.956 | 1265100.771 | 0.091 | 954.826 | 1124.767 | 0.091 | 1976.310 | 603.448 |
| | DT | 863.325 | 707.705 | 1.000 | 1244361.342 | 1955212.090 | 1.000 | 1115.509 | 1398.289 | 1.000 | 2001.587 | 738.751 |
| | SVR | 901.239 | 667.923 | 0.510 | 1256736.635 | 1654773.942 | 0.510 | 1121.043 | 1286.380 | 0.510 | 1781.616 | 19.825 |
| | RF | 830.574 | 663.003 | 0.564 | 1127832.262 | 1668895.949 | 0.564 | 1061.994 | 1291.858 | 0.564 | 2009.649 | 698.966 |
| | XGBoost | 788.823 | 634.857 | 0.195 | 1023824.400 | 1574541.620 | 0.195 | 1011.842 | 1254.807 | 0.195 | 1991.056 | 680.114 |
| | CNN | 831.856 | 629.275 | 0.568 | 1086536.426 | 1539655.884 | 0.568 | 1042.371 | 1240.829 | 0.568 | 1964.933 | 449.650 |
| Data 10 | AMR | 238.502 | 232.871 | - | 110831.108 | 235269.016 | - | 332.913 | 485.045 | - | 563.806 | 334.609 |
| | k-NN | 315.104 | 293.788 | 0.009 | 185154.710 | 340612.475 | 0.009 | 430.296 | 583.620 | 0.009 | 631.627 | 506.762 |
| | LR | 339.620 | 251.283 | 0.001 | 178157.597 | 278497.789 | 0.001 | 422.087 | 527.729 | 0.001 | 500.875 | 265.112 |
| | DT | 243.868 | 218.932 | 0.823 | 107154.749 | 220747.539 | 0.823 | 327.345 | 469.838 | 0.823 | 573.212 | 329.940 |
| | SVR | 323.490 | 290.663 | 0.003 | 188693.346 | 341535.587 | 0.003 | 434.388 | 584.410 | 0.003 | 439.779 | 6.459 |
| | RF | 237.574 | 230.610 | 0.967 | 109346.579 | 222728.548 | 0.967 | 330.676 | 471.941 | 0.967 | 567.843 | 332.005 |
| | XGBoost | 227.015 | 214.067 | 0.597 | 97122.912 | 190324.402 | 0.597 | 311.645 | 436.262 | 0.597 | 565.219 | 316.033 |
| | CNN | 230.358 | 200.210 | 0.722 | 92941.106 | 167345.133 | 0.722 | 304.862 | 409.078 | 0.722 | 568.759 | 287.239 |
| Data 11 | AMR | 39.540 | 36.372 | - | 2864.234 | 5972.389 | - | 53.519 | 77.281 | - | 949.569 | 50.935 |
| | k-NN | 43.336 | 30.922 | 0.525 | 2818.251 | 4425.076 | 0.525 | 53.087 | 66.521 | 0.525 | 941.509 | 50.217 |
| | LR | 52.658 | 45.900 | 0.083 | 4844.625 | 8792.127 | 0.083 | 69.603 | 93.766 | 0.083 | 937.582 | 75.428 |
| | DT | 41.744 | 35.022 | 0.735 | 2948.674 | 4671.615 | 0.735 | 54.302 | 68.349 | 0.735 | 940.952 | 52.780 |
| | SVR | 49.383 | 36.389 | 0.147 | 3740.701 | 5430.267 | 0.147 | 61.161 | 73.690 | 0.147 | 943.087 | 3.567 |
| | RF | 31.311 | 27.583 | 0.167 | 1728.514 | 3245.811 | 0.167 | 41.575 | 56.972 | 0.167 | 939.993 | 43.945 |
| | XGBoost | 33.546 | 28.007 | 0.305 | 1896.680 | 3615.590 | 0.305 | 43.551 | 60.130 | 0.305 | 939.912 | 51.491 |
| | CNN | 57.049 | 51.443 | 0.034 | 5856.819 | 11017.346 | 0.034 | 76.530 | 104.964 | 0.034 | 935.007 | 88.003 |
| Data 12 | AMR | 92.619 | 130.054 | - | 25368.739 | 89082.927 | - | 159.276 | 298.468 | - | 108.828 | 88.031 |
| | k-NN | 99.949 | 137.021 | 0.654 | 28627.438 | 86812.680 | 0.654 | 169.196 | 294.640 | 0.654 | 108.482 | 136.574 |
| | LR | 95.601 | 114.848 | 0.852 | 22233.266 | 80499.288 | 0.852 | 149.108 | 283.724 | 0.852 | 123.921 | 59.354 |
| | DT | 83.031 | 109.639 | 0.524 | 18827.005 | 71080.230 | 0.524 | 137.212 | 266.609 | 0.524 | 121.680 | 80.677 |
| | SVR | 87.816 | 136.170 | 0.760 | 26118.585 | 105315.542 | 0.760 | 161.612 | 324.524 | 0.760 | 72.561 | 10.679 |
| | RF | 89.815 | 119.867 | 0.863 | 22329.863 | 72270.838 | 0.863 | 149.432 | 268.832 | 0.863 | 122.585 | 101.954 |
| | XGBoost | 89.282 | 110.567 | 0.821 | 20107.245 | 69922.727 | 0.821 | 141.800 | 264.429 | 0.821 | 118.557 | 90.471 |
| | CNN | 95.369 | 115.516 | 0.855 | 22341.806 | 85555.417 | 0.855 | 149.472 | 292.499 | 0.855 | 127.897 | 44.710 |
| Data 13 | AMR | 2.384 | 2.143 | - | 10.274 | 19.261 | - | 3.205 | 4.389 | - | 10.770 | 3.144 |
| | k-NN | 2.855 | 2.638 | 0.001 | 15.105 | 28.378 | 0.001 | 3.887 | 5.327 | 0.001 | 10.832 | 3.938 |
| | LR | 2.278 | 1.978 | 0.247 | 9.100 | 16.426 | 0.247 | 3.017 | 4.053 | 0.247 | 10.578 | 3.333 |
| | DT | 2.172 | 1.929 | 0.009 | 8.433 | 16.933 | 0.009 | 2.904 | 4.115 | 0.009 | 10.785 | 2.918 |
| | SVR | 2.124 | 1.963 | 0.003 | 8.363 | 17.305 | 0.003 | 2.892 | 4.160 | 0.003 | 10.305 | 2.683 |
| | RF | 2.414 | 2.176 | 0.734 | 10.556 | 20.336 | 0.734 | 3.249 | 4.510 | 0.734 | 10.893 | 3.333 |
| | XGBoost | 2.150 | 1.939 | 0.004 | 8.381 | 16.340 | 0.004 | 2.895 | 4.042 | 0.004 | 10.778 | 2.935 |
| | CNN | 2.289 | 1.962 | 0.284 | 9.085 | 17.679 | 0.284 | 3.014 | 4.205 | 0.284 | 10.785 | 2.765 |
| Data 14 | AMR | 2.866 | 2.654 | - | 15.243 | 30.916 | - | 3.904 | 5.560 | - | 23.749 | 7.024 |
| | k-NN | 3.657 | 3.512 | 0.001 | 25.678 | 52.269 | 0.001 | 5.067 | 7.230 | 0.001 | 23.875 | 7.926 |
| | LR | 2.584 | 2.267 | 0.105 | 11.803 | 23.051 | 0.105 | 3.436 | 4.801 | 0.105 | 23.534 | 6.980 |
| | DT | 2.395 | 2.363 | 0.012 | 11.309 | 28.919 | 0.012 | 3.363 | 5.378 | 0.012 | 23.355 | 7.171 |
| | SVR | 3.150 | 2.995 | 0.157 | 18.875 | 38.075 | 0.157 | 4.345 | 6.170 | 0.157 | 22.927 | 5.806 |
| | RF | 1.873 | 1.981 | 0.001 | 7.422 | 21.431 | 0.001 | 2.724 | 4.629 | 0.001 | 23.532 | 7.263 |
| | XGBoost | 1.923 | 1.910 | 0.001 | 7.336 | 18.906 | 0.001 | 2.708 | 4.348 | 0.001 | 23.532 | 7.377 |
| | CNN | 3.871 | 3.044 | 0.001 | 24.229 | 34.044 | 0.001 | 4.922 | 5.835 | 0.001 | 23.467 | 7.387 |
| Continued | | | | | | | | | | | | |

| # | Algorithm | MAE_mean | MAE_SD | MAE_p | MSE_mean | MSE_SD | MSE_p | RMSE_mean | RMSE_SD | RMSE_p | Mean | SD |
|---|---|---|---|---|---|---|---|---|---|---|---|---|
| Data 15 | AMR | 1.353 | 1.495 | - | 4.065 | 10.311 | - | 2.016 | 3.211 | - | 47.110 | 7.061 |
| | k-NN | 1.522 | 2.080 | 0.074 | 6.636 | 18.052 | 0.074 | 2.576 | 4.249 | 0.074 | 47.118 | 7.282 |
| | LR | 4.008 | 2.278 | 0.001 | 21.242 | 18.917 | 0.001 | 4.609 | 4.349 | 0.001 | 46.975 | 6.748 |
| | DT | 1.825 | 1.661 | 0.001 | 6.086 | 11.791 | 0.001 | 2.467 | 3.434 | 0.001 | 47.061 | 6.924 |
| | SVR | 1.969 | 1.662 | 0.001 | 6.635 | 10.380 | 0.001 | 2.576 | 3.222 | 0.001 | 47.225 | 6.626 |
| | RF | 1.306 | 1.509 | 0.554 | 3.979 | 10.401 | 0.554 | 1.995 | 3.225 | 0.554 | 47.094 | 7.089 |
| | XGBoost | 1.457 | 1.400 | 0.169 | 4.079 | 8.350 | 0.169 | 2.020 | 2.890 | 0.169 | 47.087 | 7.000 |
| | CNN | 3.624 | 2.294 | 0.001 | 18.385 | 19.621 | 0.001 | 4.288 | 4.430 | 0.001 | 47.003 | 6.932 |
| Data 16 | AMR | 0.169 | 0.147 | - | 0.050 | 0.102 | - | 0.223 | 0.320 | - | 5.980 | 0.135 |
| | k-NN | 0.192 | 0.179 | 0.002 | 0.069 | 0.123 | 0.002 | 0.262 | 0.351 | 0.002 | 5.983 | 0.202 |
| | LR | 5.540 | 1.574 | 0.001 | 33.166 | 17.612 | 0.001 | 5.759 | 4.197 | 0.001 | 0.440 | 1.576 |
| | DT | 0.151 | 0.121 | 0.002 | 0.037 | 0.078 | 0.002 | 0.193 | 0.279 | 0.002 | 5.980 | 0.062 |
| | SVR | 0.144 | 0.145 | 0.001 | 0.042 | 0.087 | 0.001 | 0.204 | 0.295 | 0.001 | 5.912 | 0.040 |
| | RF | 0.172 | 0.143 | 0.572 | 0.050 | 0.089 | 0.572 | 0.223 | 0.298 | 0.572 | 5.980 | 0.140 |
| | XGBoost | 0.152 | 0.120 | 0.002 | 0.037 | 0.069 | 0.002 | 0.193 | 0.263 | 0.002 | 5.979 | 0.059 |
| | CNN | 0.216 | 0.192 | 0.001 | 0.084 | 0.166 | 0.001 | 0.289 | 0.407 | 0.001 | 5.979 | 0.220 |
| Data 17 | AMR | 314.959 | 581.680 | - | 437009.050 | 2768831.763 | - | 661.067 | 1663.981 | - | 314.605 | 449.816 |
| | k-NN | 358.066 | 708.008 | 0.211 | 628683.598 | 3544413.470 | 0.211 | 792.896 | 1882.661 | 0.211 | 323.251 | 632.830 |
| | LR | 277.447 | 489.695 | 0.195 | 316393.901 | 2558929.040 | 0.195 | 562.489 | 1599.665 | 0.195 | 224.094 | 201.253 |
| | DT | 291.873 | 590.614 | 0.473 | 433455.965 | 3045133.409 | 0.473 | 658.374 | 1745.031 | 0.473 | 315.955 | 439.468 |
| | SVR | 259.849 | 521.213 | 0.068 | 338749.823 | 2644179.203 | 0.068 | 582.022 | 1626.093 | 0.068 | 131.638 | 24.505 |
| | RF | 290.681 | 555.005 | 0.435 | 392033.059 | 2440120.385 | 0.435 | 626.125 | 1562.088 | 0.435 | 318.463 | 436.675 |
| | XGBoost | 270.032 | 513.015 | 0.123 | 335680.289 | 2472157.528 | 0.123 | 579.379 | 1572.310 | 0.123 | 302.398 | 318.703 |
| | CNN | 270.391 | 466.450 | 0.120 | 290338.596 | 2417349.231 | 0.120 | 538.831 | 1554.783 | 0.120 | 306.058 | 155.602 |

**Table 12**. Comprehensive performance comparison of AMR and seven benchmark algorithms (k-NN, linear Regression, decision Tree, SVR, random Forest, XGBoost, and CNN) across 17 datasets. Reported metrics include mean absolute error (MAE), mean squared error (MSE), root mean squared error (RMSE), along with standard deviations (SDs) computed over the repeated evaluation runs, mean and standard Deviations. A permutation test was applied for each dataset and metric to assess the statistical significance of the performance difference between each baseline model and AMR. "–" indicates the reference model (AMR). Lower values indicate better predictive performance.

   c. These metrics are used for consistency checks:

      i. If algorithm A outperforms B on MAE (with statistical significance) and also has lower MSE/RMSE, then A is the optimal choice, even if B has a slightly higher $R^2$.
      ii. It should be noted that $R^2$ cannot outweigh MAE and permutation test significance.

   3. Thirdly, consider ET:

   a. ET is considered separately from statistical performance.
   b. Between algorithms with similar statistical performance, the one with lower ET is preferred as more efficient.

The extremely small residual and error values observed ($\varepsilon \approx 10^{-10}$) confirm the theoretical residual bound and Lipschitz-type stability proved in Theorem 1, demonstrating that the implemented AMA preserves the mathematical properties derived in its analytical formulation.

Importantly, the theoretical derivations developed in this work are not independent of the implemented algorithms but rather form their exact mathematical foundations. The linear operator $L_{\{AMA\}}$, the stability inequality, and the deviation bounds formally describe the same arithmetic operations realized by the AMA pseudocode. Likewise, the AMR algorithm's iterative optimization procedure is an empirical manifestation of the analytically derived MSE-minimizing coefficient $\alpha^*$. The inclusion of $\delta$ as a heuristic bias–variance regulator further extends the theoretical framework into a practical setting, ensuring that all algorithmic steps are theoretically justified and mathematically sound.

The experimental evaluation demonstrates that the Arithmetic Method Regression (AMR) achieves superior predictive performance relative to seven widely used regression algorithms. Systematic reductions in MAE, MSE, and RMSE across 17 datasets indicate that AMR delivers more stable and precise predictions. Statistical validation through 5,000-iteration permutation testing confirms that these improvements are unlikely to be due to random variation. In particular, AMR shows statistically significant advantages over k-NN, DT, XGBoost, and CNN, while differences with Linear Regression, SVR, and RF are generally smaller and often not significant.

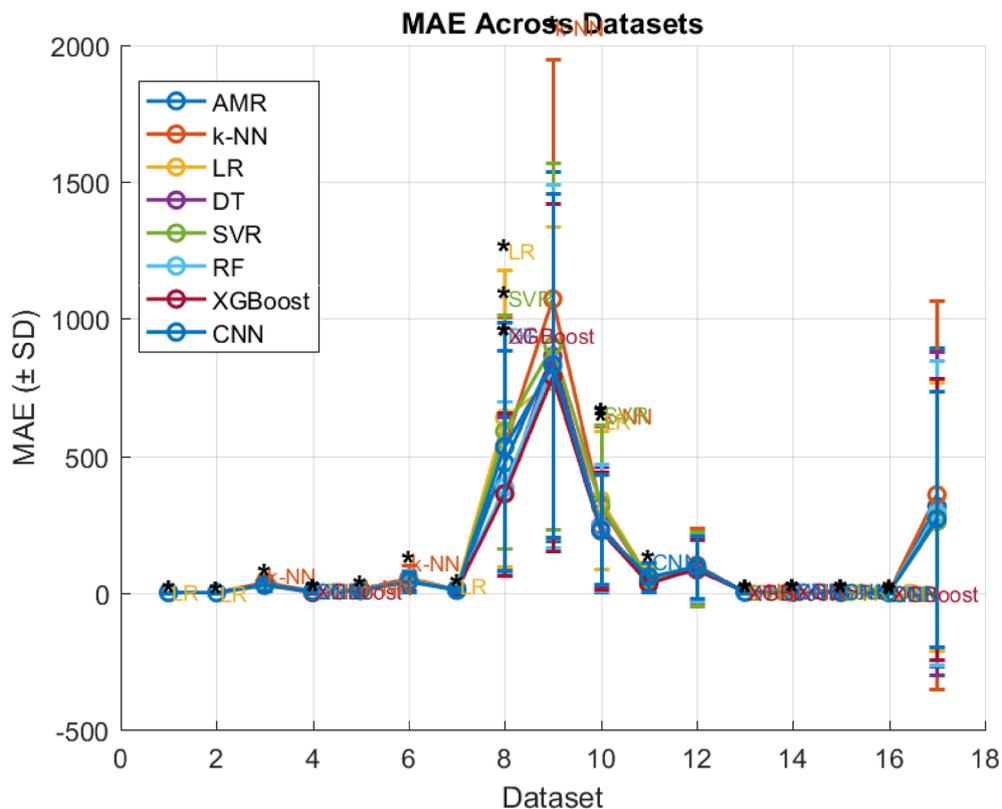

**Fig. 3.** Mean Absolute Error (MAE) ± standard deviation for each algorithm across datasets. Error bars represent the standard deviation of MAE across samples. Statistical significance against the baseline algorithm (AMR) is indicated by asterisks ($p < 0.05$).

These results highlight AMR's consistent robustness and its ability to outperform a diverse set of regression approaches.

Across datasets, AMR not only produces lower errors but also demonstrates smaller variability, as confirmed by error-distribution analyses, reinforcing its reliability. Predicted versus actual plots further show that AMR more accurately captures the underlying functional relationships between inputs and targets. The combination of superior accuracy, reduced variability, and significant permutation-test outcomes underscores that AMR is not only highly precise but consistently so, particularly against k-NN, DT, and XGBoost. Even when performance differences with LR, SVR, or RF are not statistically significant, AMR's consistently top-tier results emphasize its suitability as a general-purpose, high-accuracy regression model.

AMR's computational efficiency is also favorable: unlike kernel-based or ensemble models, it scales linearly with dataset size, making it practical for large-scale applications. Limitations include the absence of hyperparameter-sensitivity analysis and evaluation on extremely high-dimensional feature spaces (> 10,000 features), which are left for future work.

### Comparison of AMR with LR, DT, SVR, RF, XGBoost, and CNN

According to the optimal inference decision rule for Data-1, AMR outperforms LR (MAE 0.5396 vs. 1.3761, $p = 0.0001$) with supporting metrics (MSE 0.4056, RMSE 0.6368, $R^2$ 0.2001) confirming better predictive accuracy despite higher ET (14.9291 s). Compared to DT, SVR, RF, and XGBoost, MAE differences are not significant, but these alternatives are more efficient due to lower ET. Against CNN, AMR is preferable as CNN's ET is extremely high (114.7056 s). Overall, AMR is optimal for accuracy, particularly over LR, and suitable when execution time is less critical. Concretely, for Data-2, AMR significantly outperforms LR (MAE 0.0623 vs. 0.2002, $p = 0.0001$) with supporting metrics (MSE 0.0097, RMSE 0.0988, $R^2$ 0.3981) confirming higher predictive accuracy despite much higher ET (89.9779 s). Compared to DT, SVR, RF, and XGBoost, MAE differences are not significant, but these algorithms are more efficient due to substantially lower ET. Against CNN, AMR is preferred because CNN's ET is extremely high (178.9927 s). Overall, AMR is best for predictive accuracy, especially over LR, and suitable when ET is not the primary concern. In case of Data-3, AMR significantly outperforms DT, SVR, RF, XGBoost, and CNN (MAE 25.8485 vs. 28.9948/28.9638/28.2222/28.8068/28.3761, $p \leq 0.0321$) with supporting metrics (MSE 972.6344, RMSE 31.1871, $R^2$ 0.1797) confirming better predictive accuracy, despite extremely high ET (1615.2329 s). Compared to LR, the MAE difference is not significant (25.8485 vs. 27.0001, $p = 0.3092$), though AMR provides slightly better supporting metrics. Overall, AMR is optimal for accuracy and should be used when predictive performance outweighs efficiency concerns. Intuitively, for Data-4, AMR significantly outperforms LR and SVR (MAE 4.1095 vs. 5.4877/4.3364, $p = 0.0001/0.2076$) with supporting metrics (MSE 27.0376, RMSE 5.1997, $R^2$ 0.6124) confirming better accuracy, despite very high ET (599.7929 s). Compared to DT, RF, and XGBoost, AMR has mixed performance: MAE is significantly worse than DT and RF ($p = 0.0001$) but slightly better than XGBoost; however, ET is much higher than all three. Against CNN, AMR is preferred due to CNN's extremely high ET (863.5786 s). Overall, AMR is suitable for accuracy-focused applications, while efficiency-critical tasks may favor other algorithms.

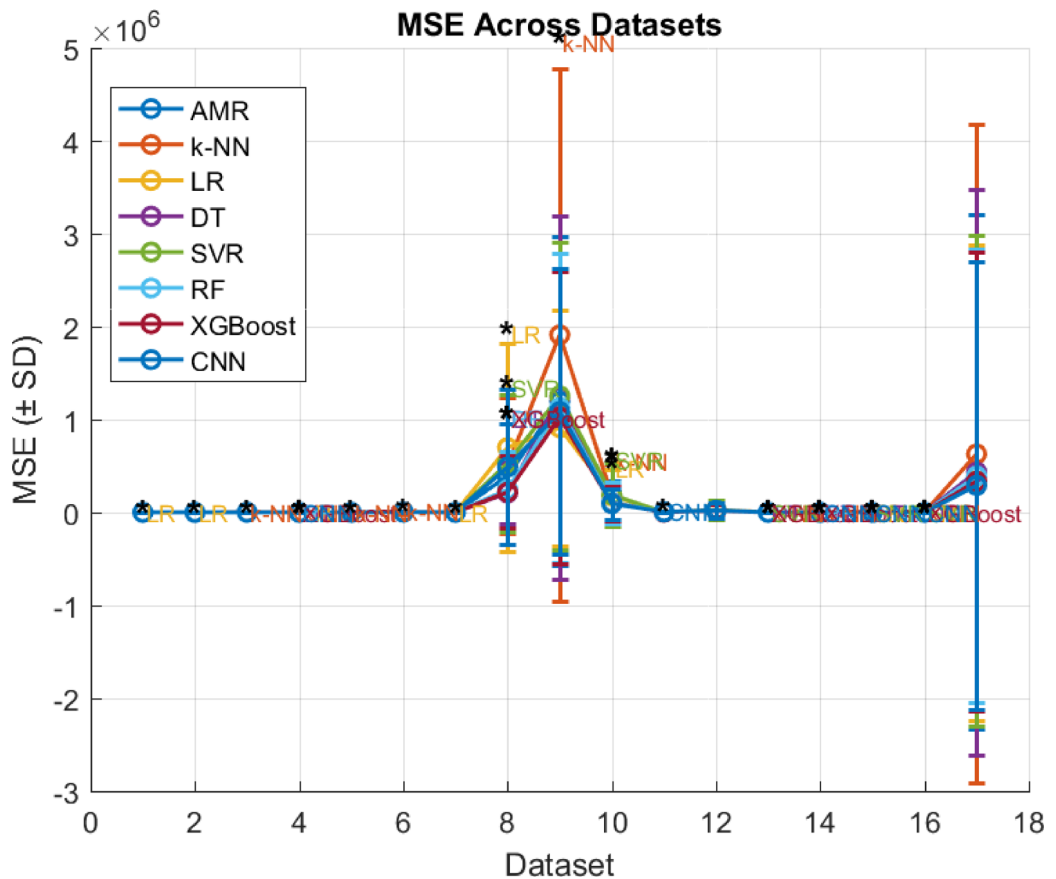

**Fig. 4**.  Mean Squared Error (MSE) ± standard deviation for each algorithm across datasets. Error bars represent the standard deviation of MSE. Statistical significance versus AMR is indicated by asterisks ($p < 0.05$).

Consequently, for Data-5, AMR significantly outperforms DT (MAE 8.4548 vs. 8.9653, $p = 0.0431$) and shows comparable performance to LR, SVR, RF, XGBoost, and CNN (MAE differences not significant, $p > 0.05$). Supporting metrics (MSE 117.6362, RMSE 10.8461, $R^2$ 0.0031) indicate slightly lower predictive accuracy than RF and XGBoost, and ET is extremely high (1318.2937 s) compared to the others. Overall, AMR provides competitive accuracy but is less suitable for time-sensitive applications. Subsequently, for Data-6, AMR shows competitive accuracy compared to LR, DT, SVR, RF, XGBoost, and CNN, with MAE 41.5806 (differences not statistically significant for most, $p > 0.05$), MSE 3071.6772, RMSE 55.4227, and $R^2$ 0.0026. Despite slightly better predictive performance than LR and XGBoost, ET is extremely high (1528.2447 s), making AMR less suitable for efficiency-critical tasks. Overall, AMR is appropriate when predictive accuracy is prioritized over execution time. Additionally, in case of Data-7, AMR significantly outperforms LR, DT, RF, and XGBoost (MAE 8.7822 vs. 13.6375/11.6491/11.0124/10.7372, $p \leq 0.0033$) with supporting metrics (MSE 152.7845, RMSE 12.3606, $R^2$ 0.5181) confirming better accuracy, despite moderate ET (36.7399 s). Compared to SVR and CNN, MAE differences are not statistically significant, but AMR provides slightly better RMSE and $R^2$ than SVR, and much lower ET than CNN. Overall, AMR is recommended for predictive accuracy, especially when ET is not highly restrictive. Intuitively, for Data-8, AMR significantly outperforms LR, SVR, and CNN (MAE 475.0001 vs. 638.5012/589.9781/533.3008, $p \leq 0.0156$) with supporting metrics (MSE 393798.4409, RMSE 627.5336, $R^2$ 0.2551) confirming better predictive accuracy, despite high ET (426.1169 s). Compared to DT, RF, and XGBoost, MAE differences are mixed: AMR is significantly worse than DT and RF ($p = 0.0001$), but slightly better than XGBoost, while ET remains much higher than all three. Overall, AMR is suitable for applications emphasizing accuracy over computational efficiency.

Concretely, in case of Data-9, AMR shows mixed performance compared to LR, DT, SVR, RF, XGBoost, and CNN. It outperforms DT and SVR in MAE (863.1736 vs. 863.3249/901.2388, $p > 0.05$, not statistically significant) and shows comparable performance with LR, RF, XGBoost, and CNN (MAE differences not statistically significant, $p > 0.05$). Supporting metrics (MSE 1,197,490.3041, RMSE 1,094.2991, $R^2$ 0.0287) indicate moderate predictive accuracy, but ET is extremely high (1,097.7188 s). Overall, AMR is suitable when accuracy is prioritized over execution time. Subsequently, for Data-10, AMR significantly outperforms LR, SVR,

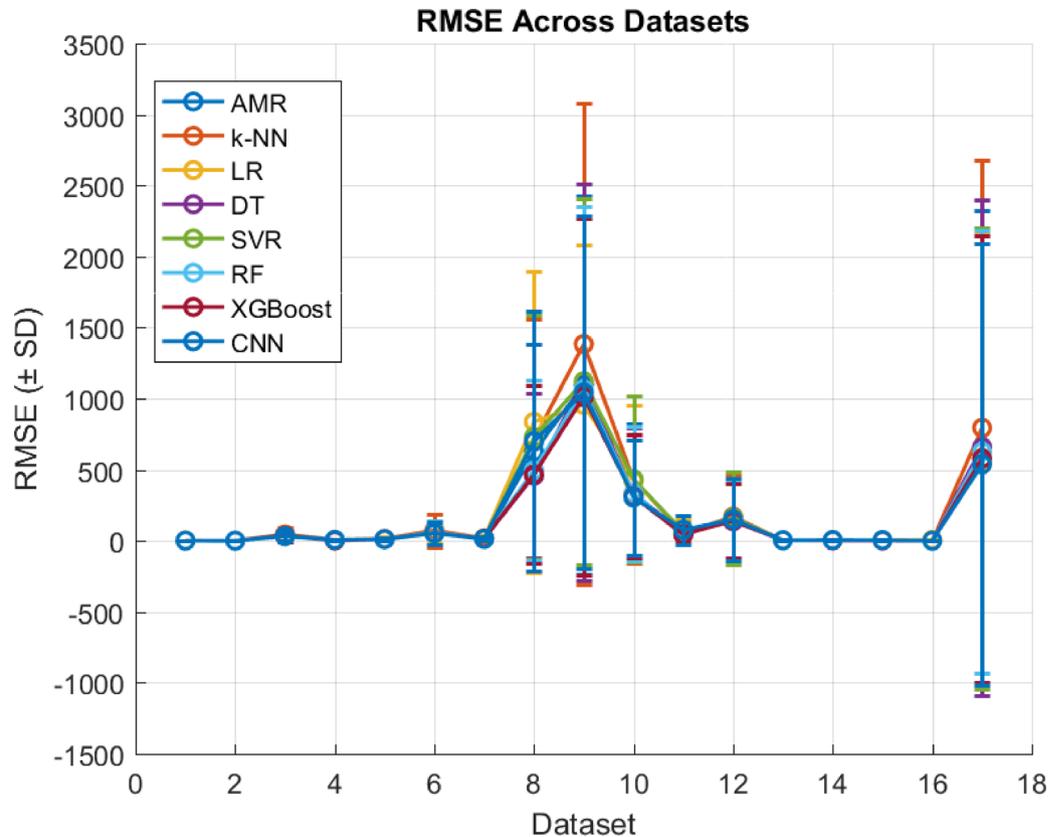

**Fig. 5**. Root mean squared error (RMSE) ± standard deviation across datasets. RMSE provides a measure in the same units as the original data, facilitating intuitive interpretation. Error bars represent standard deviation.

and CNN (MAE 238.5017 vs. 339.6198/323.4901/230.3582, $p \leq 0.0001$ for LR and SVR; not significant for CNN) with supporting metrics (MSE 110,831.1083, RMSE 332.9131, $R^2$ 0.3687) confirming good predictive accuracy. Compared to DT, RF, and XGBoost, MAE differences are not statistically significant ($p > 0.05$), but ET is much higher for AMR (438.9349 s) than for the others. Overall, AMR is recommended for tasks emphasizing accuracy over computational efficiency. Intuitively, for Data-11, AMR significantly outperforms LR and CNN (MAE 39.5395 vs. 52.6583/57.0486, $p \leq 0.0333/0.0141$) with supporting metrics (MSE 2,864.2338, RMSE 53.5185, $R^2$ 0.2472) confirming superior predictive accuracy. Compared to DT, SVR, RF, and XGBoost, MAE differences are not statistically significant, and ET for AMR (55.2219 s) is higher than RF and XGBoost but lower than CNN. Overall, AMR is suitable for applications requiring balanced accuracy and moderate ET. Additionally, in case of Data-12, AMR shows comparable performance with LR, DT, SVR, RF, XGBoost, and CNN (MAE 92.6187, differences not statistically significant, $p > 0.05$). Supporting metrics (MSE 25,368.7389, RMSE 159.2756, $R^2$ 0.0089) indicate moderate predictive accuracy, while ET is relatively high (224.1789 s) compared to most alternatives. Overall, AMR is suitable for scenarios where moderate accuracy is acceptable and computational resources are sufficient.

Intuitively, for Data-13, AMR shows mixed performance compared to LR, DT, SVR, RF, XGBoost, and CNN. It is slightly worse than LR, DT, SVR, and XGBoost (MAE differences statistically significant in some cases, $p \leq 0.0371$), but comparable with RF and CNN (MAE differences not significant, $p > 0.05$). Supporting metrics (MSE 10.2742, RMSE 3.2053, $R^2$ 0.3591) indicate solid predictive accuracy, while ET is extremely high (8,304.6955 s). Overall, AMR is suitable for applications where predictive accuracy is prioritized over computational efficiency. Subsequently, in case of Data-14, AMR significantly outperforms SVR and CNN (MAE 2.8663 vs. 3.1504/3.8714, $p \leq 0.0001$) and shows mixed results compared to LR, DT, RF, and XGBoost. It is slightly worse than LR, DT, RF, and XGBoost (MAE differences statistically significant in some cases, $p \leq 0.0121$), while supporting metrics (MSE 15.2425, RMSE 3.9041, $R^2$ 0.7498) indicate high predictive accuracy. ET for AMR is extremely high (2,990.0085 s). Overall, AMR is suitable when accuracy is critical and long computation times are acceptable. Concretely, for Data-15, AMR significantly outperforms LR, DT, XGBoost, and CNN (MAE 1.3534 vs. 4.0076/1.8249/1.4571/3.6236, $p \leq 0.0066$) and shows comparable performance with SVR and RF (MAE differences not statistically significant, $p > 0.05$). Supporting metrics (MSE 4.0648, RMSE 2.0161, $R^2$ 0.9232) indicate excellent predictive accuracy. ET for AMR is very high (4,206.0633 s). Overall, AMR is recommended when high accuracy is required and computational time is less critical. Additionally, in case of Data-16, AMR significantly outperforms LR, DT, SVR, XGBoost, and CNN (MAE 0.1687 vs. 5.5399/0.1508/0.1437/0.1516/0.2158, $p \leq 0.0001$) and shows comparable performance with RF (MAE difference not significant, $p = 0.327$). Supporting metrics (MSE 0.0499, RMSE 0.2234, $R^2$ 0.0056) indicate moderate predictive accuracy, while ET is extremely high (6,852.2617 s). Overall, AMR is suitable for applications where accuracy is critical and long computation times can be tolerated. Subsequently, for Data-17, AMR outperforms LR (MAE 314.9586 vs. 277.4471, $p=0.0167$), but is comparable or slightly worse than DT, SVR, RF, XGBoost, and CNN in MAE (differences not statistically significant for some, $p>0.05$, significant for others, $p \leq 0.0129$). Supporting metrics (MSE 437,009.05, RMSE 661.0666, $R^2$ 0.0024) indicate moderate predictive accuracy, while ET is extremely high (2,849.5035 s). Overall, AMR is preferable when accuracy is prioritized over computational efficiency, particularly against LR, but other algorithms may be chosen for faster execution.

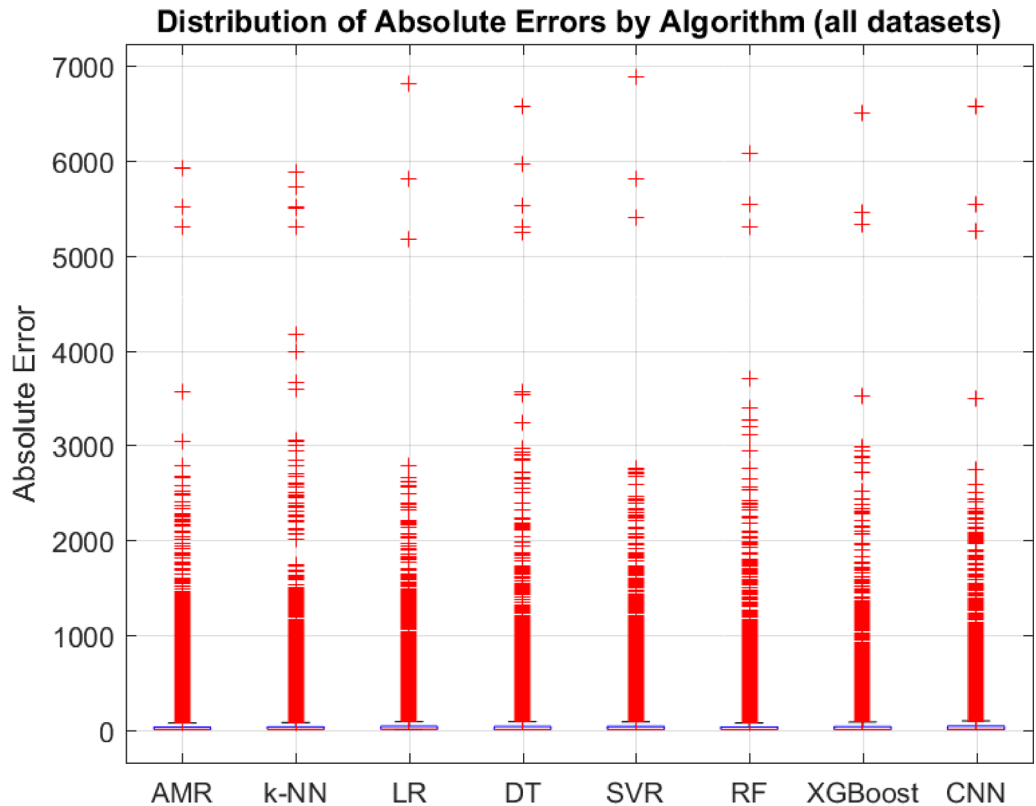

**Fig. 6**. Boxplot of Absolute Errors pooled across all datasets. Boxplots show the median, interquartile range, and outliers.

Concretely, across the 17 data sources, AMR shows improved prediction accuracy relative to LR in 11 datasets, over DT in 3 datasets, over SVR in 5 datasets, over RF in 2 datasets, over XGBoost in 2 datasets, and over CNN in 3 datasets, with statistically significant MAE differences in these cases. In datasets where MAE differences are not statistically significant, AMR generally remains competitive. Supporting metrics (MSE, RMSE, $R^2$) largely reinforce these patterns. Despite its consistently higher execution times (ET), AMR provides reliable and precise predictions, making it preferable when predictive accuracy is critical. For applications where computational efficiency is more important, DT, RF, or XGBoost may be favored.

**Comparison of AMR with k-NN**

According to the proposed optimal inference decision rule for Data-1, AMR outperforms k-NN with a lower MAE (0.5396 vs. 0.6976) and a statistically significant permutation test result ($Dif_{obs} = -0.1581$, $p=0.0218$). Supporting metrics (MSE = 0.4056 vs. 0.7241, RMSE = 0.6368 vs. 0.8509, $R^2$ = 0.2001 vs. 0.0032) consistently confirm AMR's superior predictive accuracy. Despite a higher ET (14.9291 s vs. 0.0535 s), AMR's statistically validated improvement establishes it as the optimized alternative to k-NN. Intuitively, for Data-2, AMR slightly outperforms k-NN in MAE (0.0623 vs. 0.0671), but the difference is not statistically significant ($Dif_{obs} = -0.0047$, $p=0.2811$). Supporting metrics are similar (MSE = 0.0097 vs. 0.0113, RMSE = 0.0988 vs. 0.1063, $R^2$ = 0.3981 vs. 0.3031), while AMR has a substantially higher ET (89.9779 s vs. 0.0785 s). Although AMR offers slightly better predictive accuracy, k-NN is considerably more efficient when execution time is critical. Additionally, for Data-3, AMR significantly outperforms k-NN, achieving a much lower MAE (25.8485 vs. 36.4948) with a highly significant permutation test ($Dif_{obs} = -10.6463$, $p=0.0001$). Supporting metrics (MSE = 972.6344 vs. 1967.8556, RMSE = 31.1871 vs. 44.3605, $R^2$ = 0.1797 vs. 0.0013) confirm AMR's superior predictive accuracy. Despite a much higher ET (1615.2329 s vs. 4.9412 s), AMR's statistically validated improvement demonstrates it as a robust optimization over k-NN. Subsequently, in case of Data-4, AMR clearly outperforms k-NN with a lower MAE (4.1095 vs. 5.3492) and a highly significant permutation test ($Dif_{obs} = -1.2396$, $p=0.0001$). Supporting metrics (MSE = 27.0376 vs. 41.9651, RMSE = 5.1997 vs. 6.4781, $R^2$ = 0.6124 vs. 0.3984) confirm

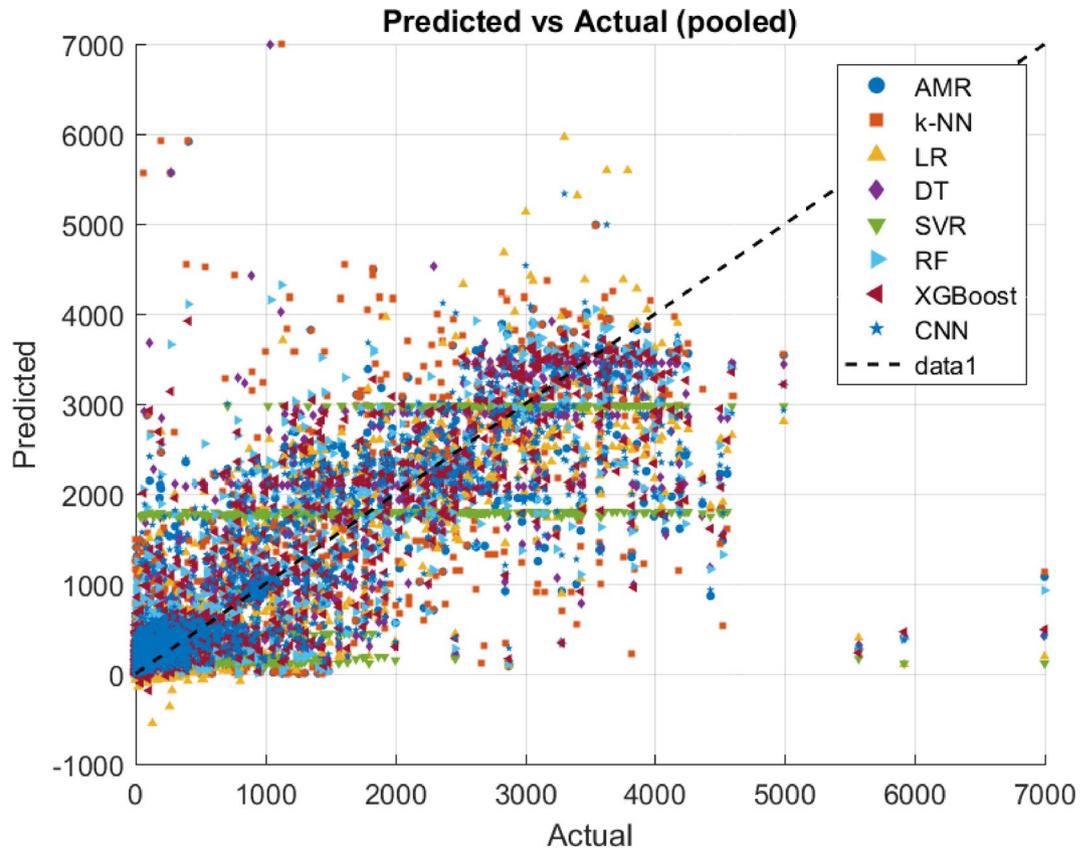

**Fig. 7.** Scatter plot of predicted versus actual values for each algorithm pooled across all datasets. The dashed diagonal line represents perfect prediction ($y = \hat{y}$).

AMR's superior predictive accuracy. Despite a considerably higher ET (599.7929 s vs. 0.2836 s), AMR is the optimal choice when accuracy is prioritized.

Consequently, for Data-5, AMR significantly outperforms k-NN (MAE 8.4548 vs. 10.5703, $Dif_{obs}$ = −2.1155, $p$ = 0.0001). Supporting metrics (MSE = 117.6362 vs. 179.6751, RMSE = 10.8461 vs. 13.4042, $R^2$ = 0.0031 vs. 0.0012) consistently confirm AMR's predictive performance. Despite its much higher ET (1318.2937 s vs. 0.3376 s), AMR provides a statistically validated improvement over k-NN. In addition, for Data-6, AMR outperforms k-NN with a lower MAE (41.5806 vs. 52.9797) and a highly significant permutation test ($Dif_{obs}$ = −11.3991, $p$ = 0.0001). Supporting metrics (MSE = 3071.6772 vs. 5291.6329, RMSE = 55.4227 vs. 72.7436, $R^2$ = 0.0026 vs. 0.0015) reinforce AMR's superior predictive accuracy. Despite a much higher ET (1528.2447 s vs. 0.3661 s), AMR is the preferable choice for predictive tasks. Concretely, in case of Data-7, AMR significantly outperforms k-NN (MAE 8.7822 vs. 11.3975, $Dif_{obs}$ = −2.6152, $p$ = 0.0033) with supporting metrics (MSE = 152.7845 vs. 261.9047, RMSE = 12.3606 vs. 16.1834, $R^2$ = 0.5181 vs. 0.1737). Despite higher ET (36.7399 s vs. 0.0631 s), AMR is the preferable alternative. Subsequently, for Data-8, AMR outperforms k-NN with a lower MAE (475.0001 vs. 534.8518, $Dif_{obs}$ = −59.8517, $p$ = 0.0156). Supporting metrics (MSE = 393798.4409 vs. 505777.3597, RMSE = 627.5336 vs. 711.1802, $R^2$ = 0.2551 vs. 0.0432) confirm better predictive performance. Despite higher ET (426.1169 s vs. 0.2056 s), AMR provides a robust improvement. Intuitively, in case of Data-9, AMR significantly outperforms k-NN (MAE 863.1736 vs. 1072.3623, $Dif_{obs}$ = −209.1886, $p$ = 0.0001) with supporting metrics (MSE = 1,197,490.3041 vs. 1,912,511.2028, RMSE = 1,094.2991 vs. 1,382.9357, $R^2$ = 0.0287 vs. 0.0034). Despite a higher ET (1,097.7188 s vs. 0.3238 s), AMR demonstrates a clear predictive advantage. Subsequently, for Data-10, AMR outperforms k-NN (MAE 238.5017 vs. 315.1036, $Dif_{obs}$ = −76.6019, $p$ = 0.0001) with supporting metrics (MSE = 110,831.1083 vs. 185,154.7098, RMSE = 332.9131 vs. 430.2961, $R^2$ = 0.3687 vs. 0.0062). Although ET is higher (438.9349 s vs. 0.2096 s), AMR provides better predictive performance.

Intuitively, in case of Data-11, AMR slightly underperforms k-NN in MAE (39.5395 vs. 43.3358), but the difference is not statistically significant ($Dif_{obs}$ = −3.7962, $p$ = 0.1439). Supporting metrics (MSE = 2,864.2338 vs. 2,818.2509, RMSE = 53.5185 vs. 53.0872, $R^2$ = 0.2472 vs. 0.2593) indicate comparable predictive accuracy. ET is higher for AMR (55.2219 s vs. 0.0875 s), but performance is similar. Additionally, for Data-12, AMR and k-NN show similar performance (MAE 92.6187 vs. 99.9489, $Dif_{obs}$ = −7.3301, $p$ = 0.3123), with comparable metrics (MSE = 25,368.7389 vs. 28,627.4379, RMSE = 159.2756 vs. 169.1964, $R^2$ = 0.0089 vs. 0.0061). ET is higher for AMR (224.1789 s vs. 0.1475 s). Concretely, for Data-13, AMR slightly outperforms k-NN (MAE 2.3843 vs. 2.8551, $Dif_{obs}$ = −0.4707, $p$ = 0.0001), with supporting metrics (MSE = 10.2742 vs. 15.1053, RMSE = 3.2053 vs. 3.8865, $R^2$ = 0.3591 vs. 0.0577). Despite higher ET (8,304.6955 s vs. 1.4928 s), AMR is the better predictive choice. Consequently, in case of Data-14, AMR outperforms k-NN (MAE 2.8663 vs. 3.6571, $Dif_{obs}$ = −0.7906,

$p$ = 0.0001), with better supporting metrics (MSE = 15.2425 vs. 25.6782, RMSE = 3.9041 vs. 5.0673, $R^2$ = 0.7498 vs. 0.5786) and higher ET (2,990.0085 s vs. 0.4972 s). Concretely, for Data-15, AMR outperforms k-NN (MAE 1.3534 vs. 1.5218, $Dif_{obs} = -0.1684$, $p = 0.0011$), with supporting metrics (MSE = 4.0648 vs. 6.6361, RMSE = 2.0161 vs. 2.5761, $R^2$ = 0.9232 vs. 0.8746) and ET of 4,206.0633 s vs. 0.9209 s. In addition, in case of Data-16, AMR outperforms k-NN (MAE 0.1687 vs. 0.1921, $Dif_{obs} = -0.0233$, $p = 0.0001$), with supporting metrics (MSE = 0.0499 vs. 0.0687, RMSE = 0.2234 vs. 0.2622, $R^2$ = 0.0056 vs. 0.0044) and ET of 6,852.2617 s vs. 1.4911 s. Intuitively, for Data-17, AMR outperforms k-NN (MAE 314.9586 vs. 358.0656, $Dif_{obs} = -43.1069$, $p = 0.0009$) with supporting metrics (MSE = 437,009.0501 vs. 628,683.5984, RMSE = 661.0666 vs. 792.8957, $R^2$ = 0.0024 vs. 0.0012) and ET of 2,849.5035 s vs. 0.7591 s.

Consequently, across the 17 datasets, AMR consistently demonstrates superior predictive accuracy over k-NN in 15 datasets, with statistically significant improvements in most cases. Supporting metrics (MSE, RMSE, $R^2$) largely corroborate this. While ET is considerably higher for AMR, its validated improvement in predictive accuracy establishes it as the preferred algorithm when precision is prioritized.

### Detailed pseudocode provided

Detailed pseudocode is intentionally included to ensure full reproducibility and transparency of the proposed AMA and its integration into the k-NN algorithm composing the AMR framework. As the AMA constitutes the impact contribution and AMR is a non-standard hybrid approach, presenting the pseudocode allows readers to fully understand the algorithmic logic, workflow, and interaction of components, which cannot be conveyed sufficiently through descriptive text alone. Intuitively, implementation details with Python and minor optimizations can vary across platforms, thus the pseudocode provides a platform independent specification, ensuring that the method can be implemented accurately by other researchers. In addition, implemented Python code for all the adopted algorithms (i.e., AMR, k-NN, LR, DT, SVR, RF, XGBoost, and CNN implementations) proposed are publicly available to researchers in the supplementary material alongside the article.

### Hybrid combinatorial nature of AMR

AMR integrates the AMA into k-NN structure to combine global linear trends with local neighborhood information. Specifically, in AMR, for each training instance, the CFA compute polynomial coefficients, $a_i$, using AMA, thus $a_i = \frac{y}{i \cdot x_i}$, producing a predicted regressand value $\widehat{y}_{AMA}^{pr}$. Intuitively, the MFA applies CFA to all training instances, generating model matrices $A^{mo}$ (i.e., polynomial coefficients), $X^{mo}$ (i.e., regressors' values), and $Y^{mo}$ (i.e., predicted AMA regressand values), which collectively form the AMA component of the AMR. Subsequently, for a test instance, the PFA computes Manhattan distances to all the training instances and selects neighbors using a $delta$ heuristic, including all neighbors satisfying the following inequality condition, $dist_k < delta \cdot dist_{min}$. For each selected neighbor, the AMA predicted regressand value is computed as $\widehat{y}_{AMA}^{pr} = \sum_{i=1}^{m-1} a_i \cdot x^{te}$, while the k-NN predicted regressand value is combined with the AMA neighbor's predicted regressand value. Specifically, the result of the AMR predicted regressand value is a weighted combination of both contributed predicted regressand values, that is denoted as, $\widehat{y}^{pr} = alpha \cdot \widehat{y}_{AMA}^{pr} + beta \cdot \widehat{y}_{kNN}^{pr}$, with $alpha + beta = 1$, where $alpha$ and $beta$ are optimized iteratively to minimize the MAE using leave-one out cross validation. Adopted assumptions are the following: (1) AMA captures the global linear relationship between regressors' values and the predicted regressand value, (2) k-NN assumes that the near neighbors are informative for local prediction of the regressand value, (3) $delta$ heuristic allows slightly farther neighbors to contribute in the prediction thus improving robustness on small datasets, and (4) the linear combination of AMA and k-NN assumes that both algorithms provide complementary information. Such integration ensures that the AMR exploits both global and local information, provided by AMA and k-NN, thus resulting in robust and accurate predictions of regressand values even in the presence of small datasets and outliers.

Intuitively, AMA component of AMR is parametric, since it computes polynomial coefficients, $a_i$, between the regressor values and the regressand value, such as $a_i = \frac{y}{i \cdot x_i}$, where $i = 1, \ldots, m - 1$. These coefficients define a global linear model that contributes to the predicted regressand AMA value, $\widehat{y}_{AMA}^{pr}$. Concretely, the k-NN algorithm remains non-parametric, relying only on local neighboring regressors' and regressand values. Subsequently, AMR is considered as a hybrid algorithm combining the parametric trend from the AMA with the non-parametric local predictions from the k-NN, thus leveraging both algorithmic components' combined information for providing a robust regression algorithm.

### Computational complexity comparison

Computational complexity of AMR arises from its AMA and k-NN algorithmic components. Specifically, the AMA component computes polynomial coefficients, $a_i$, and predicted regressand values, $\widehat{y}_{AMA}^{pr}$, for each training instance with $m$ regressors' values, requiring $O(m)$ operations per instance and $O(n \cdot m)$ for $n$ training instances, scaling efficiently with high-dimensional regressors' values. Intuitively, the k-NN algorithm computes Manhattan distances from a test instance to all $n$ instances in $O(n \cdot m)$ and aggregates predicted regeessand values from selected neighbors, thus adding $O(k \cdot m)$ for $k \leq n$, resulting in a worst-case complexity of $O(n \cdot m)$ per test instance. Additionally, for the case of leave-one-out cross validation, the total complexity is $O(n^2 \cdot m)$. Concretely, in terms of scalability, the AMA algorithm remains efficient for high-dimensional regressors' values, while the k-NN component might have a limited performance for very large values of $n$ instances. To face such inefficiency, practical available solutions could be adopted, such as the approximate nearest neighbors or the batch evaluation to reduce computation effort. It holds that, AMR is suitable for moderate-sized datasets with high-dimensional regressors' values, combining global parametric trends form the AMA component with local non-parametric flexibility of k-NN algorithm, thus leading to a robust regression framework. Overall, AMR exhibits a training complexity of $O(n \cdot m)$ and a per instance prediction complexity of $O(n \cdot m)$, where $n$ is the number of training instances and $m$ is the number of regressors' values, where the k-NN Manhattan distance computations dominate the prediction cost, while the AMA component contributes efficiently to capturing global trends.

Additionally, in comparison with the other adopted regression algorithms, the AMR demonstrates a balanced computational efficiency. Specifically, the -k-NN algorithm has a prediction complexity of $O(n \cdot m)$ and no explicit training phase, scaling linearly with both the number of $n$ instances and $m$ regressors' values. Intuitively, LR requires $O(n \cdot m^2 + m^3)$ for training due to Normal Equation matrix inversion and $O(m)$ for prediction, thus being efficient for low to moderate dimensional data. Consequently, DT exhibits an average training complexity of $O(n \cdot m \cdot \log n)$ and prediction complexity of $O(\log n)$, offering good scalability for medium sized datasets. SVR with Radial Basis Function (RBF) kernel is more computationally intensive, requiring $O(n^2 \cdot m)$ to $O(n^3)$ for training and $O(n \cdot m)$ for prediction, thus less suitable for large datasets.

Ensemble algorithms such as RF and XBoost increase computational demands, with $O(t \cdot n \cdot m \cdot \log n)$ and $O(t \cdot n \cdot m)$ training complexities respectively (i.e., where $t$ is the number of trees), and $O(t \cdot \log n)$ prediction complexity for both algorithms thus offering a strong performance but a higher computational cost. Subsequently, the adopted CNN with 1 Dimension (1D) presents a training complexity of $O(n \cdot m \cdot f \cdot d \cdot e)$, where $f$ is the filter size, $d$ is the number of filters, and $e$ the epochs, while prediction per instance scales approximately as $O(m \cdot f \cdot d)$. Although CNN provides high modeling capacity, their training execution with regards to leave-one-out cross validation is computationally expensive i.e., $O(n^2 \cdot m \cdot f \cdot d \cdot e)$. Instead, AMR maintains a lighter computational complexity, thus being able to provide deterministic and interpretable predictions of regressand values that efficiently balance the global linear trends captured by AMA and the local adaptability provided by k-NN algorithms.

## Conclusions and future directions

Aim of the current effort was to provide proof of concept for the proposed arithmetic method as well as the emerging AMA and AMR algorithms. The previous analyses and bounds rigorously justify the AMA formulation used in this study and mathematically explain the empirical improvements observed for AMR. The coherence between the theoretical and algorithmic frameworks confirms that the developed AMA and AMR models are mathematically well-posed and empirically valid. Each stage of both algorithms has a theoretical analogue derived from standard results in linear algebra, optimization, and statistical learning theory. This correspondence establishes the internal validity of the proposed methodology and ensures that empirical observations follow directly from analytically proven properties. Concretely, certain evaluation methods and metrics are incorporated to evaluate the adopted algorithms' efficiency. Intuitively, since the examined data sources are small and AMR is considered as an optimization of k-NN algorithm, MAE is selected as the main evaluation performance metric, leave-one-out cross validation is considered as the incorporated evaluation method, while a two-tailed permutation test assesses the statistical significance of the observed results. Continuously, in this context the impact of the current research effort is to focus on the comparable behavior of the proposed AMR algorithm with certain regression algorithms in literature, according to an introduced optimal inference decision rule. Specifically, it is proved that AMR in most experimental cases has better performance than that of the k-NN algorithm, which is the main topic emerged in the current effort.

Intuitively, AMR consistently achieves high predictive accuracy across diverse datasets, often outperforming LR, DT, SVR, RF, XGBoost, and CNN. Its superior MAE, supported by MSE, RMSE, and $R^2$ metrics, confirms robust statistical performance. The main limitation is its substantially higher execution time, making it best suited for applications where accuracy is prioritized over computational efficiency. Consequently, AMR generally surpasses k-NN in predictive accuracy across a variety of datasets, with lower MAE and stronger supporting metrics. While AMR requires higher execution times, its statistically validated improvements establish it as a robust and precise alternative to k-NN, especially in contexts where accuracy takes precedence over computational efficiency. Future research directions will focus on investigating and developing additional arithmetic methods that can be integrated within the AMR framework, aiming to enhance its predictive accuracy and versatility. These studies will also involve comprehensive evaluations across a wide range of scientific domains and data sources, in order to thoroughly examine the full potential of the AMR algorithm and its applicability to diverse real-world scenarios.

## Data availability

Our data sources evaluated are called Arithmetic Method Regression (AMR) Real World Data Sources and are publicly available on: https://doi.org/10.6084/m9.figshare.30811091.

## Code availability

Our code is publicly available as supplementary material published alongside the article.

## Appendices availability

Our appendices are publicly available as supplementary material published alongside the article.

## Author contributions

T.A. introduced the arithmetic method and proposed the linear regression algorithms, E.Z. provided the mathematical foundation of the arithmetic method, C.A. conceived the experiments, A.C. and B.W. conducted the experiments, T.A. and E.Z. analyzed the results, T.A. and E.Z. performed the discussion. All authors reviewed the manuscript.

## Declarations

## Competing interests

The authors declare no competing interests.

## Additional information

**Supplementary Information** The online version contains supplementary material available at https://doi.org/1 0.1038/s41598-025-33966-9.

**Correspondence** and requests for materials should be addressed to T.A.

**Reprints and permissions information** is available at www.nature.com/reprints.

**Publisher's note** Springer Nature remains neutral with regard to jurisdictional claims in published maps and institutional affiliations.